\documentclass[12pt]{article}
\pdfoutput=1   
\usepackage{hyperref}
\usepackage{times}
\usepackage{graphicx}
\usepackage{color}
\graphicspath{{figs/}{anc/}}
\newcommand{\mypara}[1]{\paragraph{#1}}
\newcommand{\citep}[1]{\cite{#1}}

\title{Can Computers Create Art?}
\author{Aaron Hertzmann \\ Adobe Research\footnote{This essay expresses my own opinions and not those of my employer. This is a preprint version of an essay to appear in the journal \textit{Arts}, Special Issue on “The Machine as Artist (for the 21st Century)” ( \url{http://www.mdpi.com/journal/arts/special_issues/Machine_Artist} ).}} 

\begin{document}
\maketitle



\begin{abstract}
This essay discusses whether computers, using Artificial Intelligence (AI), could create art. First, the history of technologies that automated aspects of art is surveyed, including photography and animation. In each case, there were initial fears and denial of the technology, followed by a blossoming of new creative and professional opportunities for artists. The current hype and reality of Artificial Intelligence (AI) tools for art making is then discussed, together with predictions about how AI tools will be used. It is then speculated about whether it could ever happen that AI systems could be credited with authorship of artwork. It is theorized that art is something created by social agents, and so computers cannot be credited with authorship of art in our current understanding. A few ways that this could change are also hypothesized.
\end{abstract}



\section{Introduction}

Artificial Intelligence (AI) research has made staggering advances recently, including many publicly-visible developments in web search, image recognition, conversational agents, and robotics. These developments have stoked fear about Artificial Intelligence's effect on many aspects of society. In the context of art, news media hype presents new image and video creation algorithms as if they are automating the creation of art. Perhaps they will empower everyday users while putting artists out of work... and, while they're at it, rob us of our humanity?


Beyond the hype, confusion about how technology influences art pervades serious discussions. Professional artists often are concerned that computers might put them out of their jobs \citep{GabeNicholas} --- a concern I've heard for decades. 
Some practitioners present their algorithms as themselves, potentially, artists \citep{PaintingFool,CAN,InnovationEngineCACM}, as do some journalists \citep{MSartist,shimon,pogue}. And, I was recently contacted by a prominent social psychologist who, inspired by recent results with neural networks, wished to conduct experiments to assess whether ordinary people might be willing to buy artwork made by a computer, and, if so, why. It was assumed that computers were already happily making their own artwork.

On the other hand, when I have informally asked friends or colleagues the question of whether computers can create art, the answer is sometimes a decisive ``No.'' Art requires human intent, inspiration, a desire to express something. Thus, by definition, there is no such thing as art created by a computer... why would anyone worry?  The concepts of art and inspiration are often spoken of in mystical terms, something special and primal beyond the realm of science and technology; as if humans can only create art because only humans have ``souls.'' Surely, there should be a more scientific explanation. 


In this essay, I tackle the question of ``Can Computers Create Art?'' This might seem like a simple question. Sometimes people ask it as a question about technological capabilities, like asking if a new car can go 100 miles per hour. For this sort of question, someone who understands the technology ought to be able to give a simple yes-or-no answer. But it is not given that any computer will ever be widely considered as an artist. To date, there is already a rich body of computer-generated art, and, in all cases, the work is credited to the human artist(s) behind the tools, such as the authors or users of the software --- and this might never change.

A more precise statement of this essay's question is: ``Could a piece of computer software ever be widely credited as the author of an artwork? What would this require?''  This is a question of the psychology and philosophy of art; I do not describe the existing technology in much detail. I also discuss the related question of ``Will AI put artists out of jobs?'' and, more generally, of whether AI will be beneficial to art and artists.


Before directly addressing these questions, I discuss the history and current state of automation for art. I begin with some historical perspective: previous moments in history when new technologies automated image and film creation, particularly the invention of photography. In each case, we see that these new technologies caused fears of displacing artists, when, in fact, the new technology both created new opportunities for artists while invigorating traditional media.

I argue that new technologies benefit art and artists, creating new tools and modes of expression, and new styles of expression. Sadly, art and science are often viewed as being separate, or, even, in opposition \citep{TwoCultures}. Yet, technological development stimulates so much of the continued vitality of art, and new artistic technologies create new job opportunities. 
Our new AI technologies follow this trend, and will for the forseeable future: new AI algorithms will provide new tools for expression and transform our art and culture in positive ways, just as so many other technologies have in the past.

\textit{Computers do not create art, people using computers create art.}  Despite many decades of procedural and computer-generated art, there has never been a computer widely accepted as the author of an artwork. To date, all ``computer-generated art'' is the result of human invention, software development, tweaking, and other kinds of direct control and authorship. We credit the human artist as author, acknowledging that the human is always the mastermind behind the work, and that the computer is a simple tool.

I then discuss whether this could ever change: would we ever agree to assign authorship to a computer?  I argue that artistic creation is primarily a \textit{social} act, an action that people primarily perform as an interaction with other humans in society. This implies that computers cannot create art, just as people do not give gifts to their coffeemakers or marry their cars. Conversely, any human can make art, because we humans are social creatures. However, the boundaries of art are fluid, and, someday, better AI could come to be viewed as true social agents. I discuss this and other scenarios where AI algorithms could come to be accepted as artists, along with some of the dangers of too eagerly accepting software as artists.

%

This essay expresses my point of view, as someone who has developed various kinds of technology for art, has followed the development of new technologies affecting art and culture over the years, and, in a few small instances, exhibited my own work as artworks. I have written it in a style that is meant to be accessible to a general audience, since these are topics that many kinds of people care about. 

\section{History and Current Practice}

Before making general claims about how computer tools relate to art, I begin by first describing several useful historical examples, including the inventions of photography, film, and computer animation.  These examples show previous situations in which a new technology appeared poised to displace artists, but, in reality, provided new opportunities and roles for artists. 
I also discuss how procedural art and computer art employs automation. In so many of these fields, outsiders view the computer or the technology as doing the work of creativity. Yet, in each of these cases, the human artist or artists behind the work are the true authors of the work.

I argue that, throughout history, technology has expanded creative and professional opportunities for artists dramatically, by providing newer and more powerful tools for artists. The advent of new technologies often causes fears of displacement among traditional artists. In fact, these new tools ultimately enable new artistic styles and inject vitality into art forms that might otherwise grow stale. These new tools also make art more accessible to wider sections of society, both as creators and as consumers. These trends are particularly visible in the past two centuries since the Industrial Revolution.  

\subsection{How photography became an artform}
\label{sec:photography}

\textit{``From today, painting is dead!'' --- Paul Delaroche, painter, at a demonstration of the daguerreotype in 1839}\footnote{The historical information from this section is distilled from two texts: Scharf \cite{ArtPhotography} and Rosenblum \cite{Rosenblum}.}

For lessons from the past about AI and art, perhaps no invention is more significant than photography.\footnote{This connection has been made previously \citep{HertzmannThesis,Blaise}, but I explore it much more thoroughly here.} Prior to the invention of photography, realistic images of the world could only be produced by artists. In today's world, we are so swamped with images that it is hard to imagine just how special and unique it must have felt to see a skillfully-executed realistic painting. The technical skills of realism were inseparable from other creative aspects.  This changed when photography automated the task of producing images of the real world. 

In 1839, the first two commercially-practical photographic processes were invented: 
Louis-Jacques-Mand\'{e} Daguerre's dagguereotype, and William Henry Fox Talbot's negative-positive process.
They were mainly presented as ways to produce practical records of the world.  
Of the two, the daguerreotype was more popular for several decades, because Talbot's process was restricted by patents. Improvements to Talbot's method eventually made the daguerreotype obsolete and evolved into modern film processes.

\begin{figure}
    \centering
    (a) \includegraphics[width=2in]{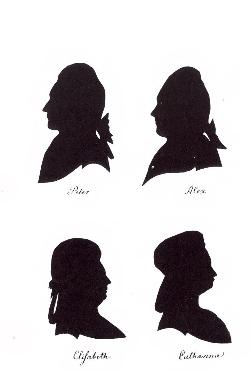}
    (b) \includegraphics[width=2in]{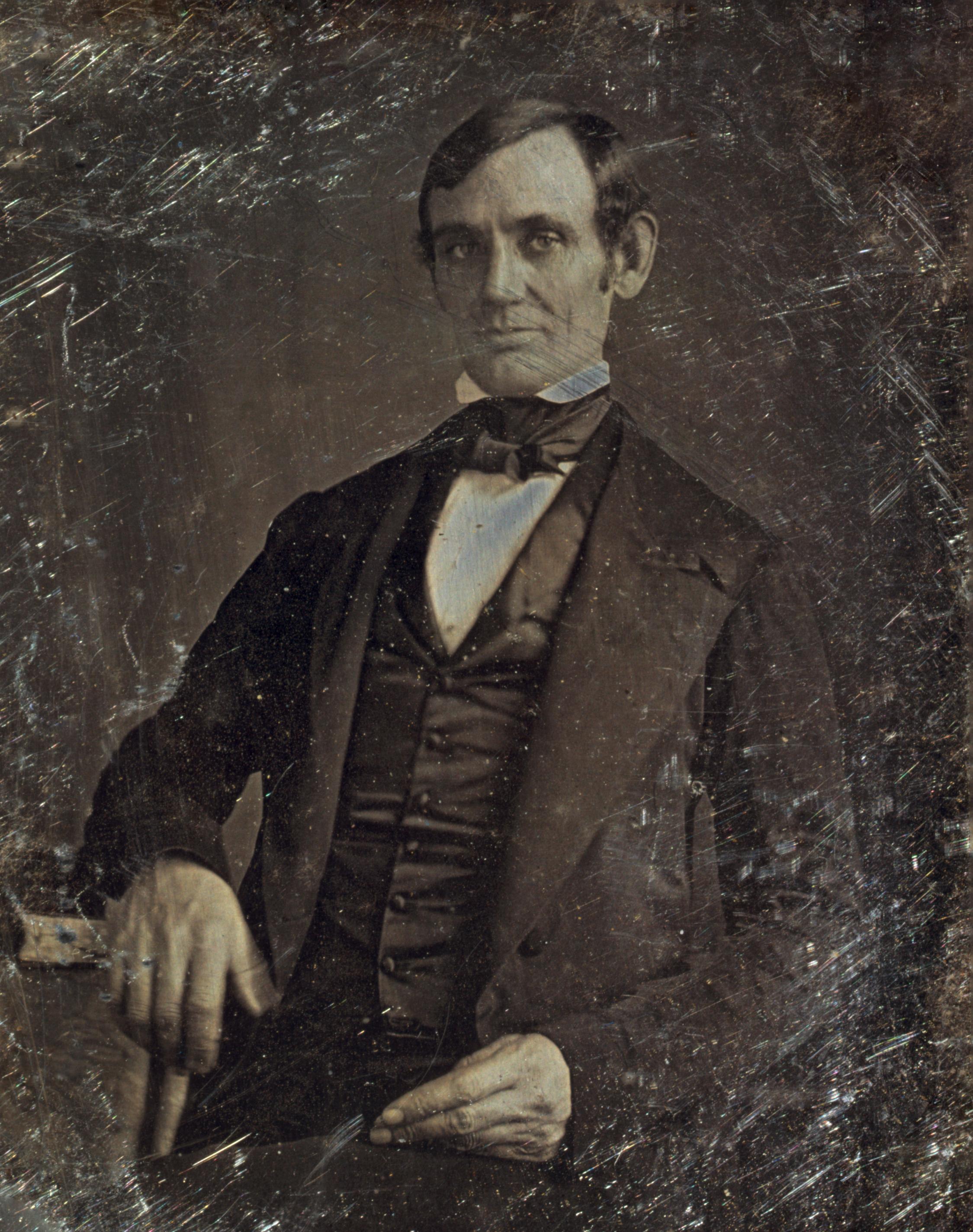}
    \caption{\label{fig:daguerreotype}
 (a) A traditional silhouette portrait.
    (b) Daguerreotype portrait of Abraham Lincoln, 1846. Photographic techniques like this completely displaced previous portraiture techniques. 
    }
\end{figure}

Portraiture was a main driver for early adoption. Then, as today, people enjoyed possessing pictures of their friends, loved-ones, and ancestors. Portrait painting was only available to aristocrats and the very wealthy. In the 18th century, several  inexpensive alternatives were developed, such as the silhouette, a representation of an individual's outline (Figure \ref{fig:daguerreotype}(a)), typically hand-cut by an artisan out of black paper. The daguerreotype offered an economical way to create a realistic portrait (Figure \ref{fig:daguerreotype}(b)). It was very slow and required locking the subject's head in place with a head brace for several minutes, while the subject tightly gripped their chair, so as not to move their fingers. Nonetheless, numerous daguerreotype studios arose and became commonplace as technologies improved, and many portraitists switched to this new technology. 
By 1863, a painter-photographer named Henri Le Secq said ``One knows that photography has harmed painting considerably, and has killed portraiture especially, once the livelihood of the artist.''
Photography  largely replaced most older forms of portraiture, such as the silhouette, and no one seems to particularly regret this loss. As much as I appreciate the mystery and beauty at looking at old etchings and portraits, I'd rather use my mobile phone camera for my own pictures than to try to paint them by hand.

\begin{figure}
    \centering
    \includegraphics[width=3in]{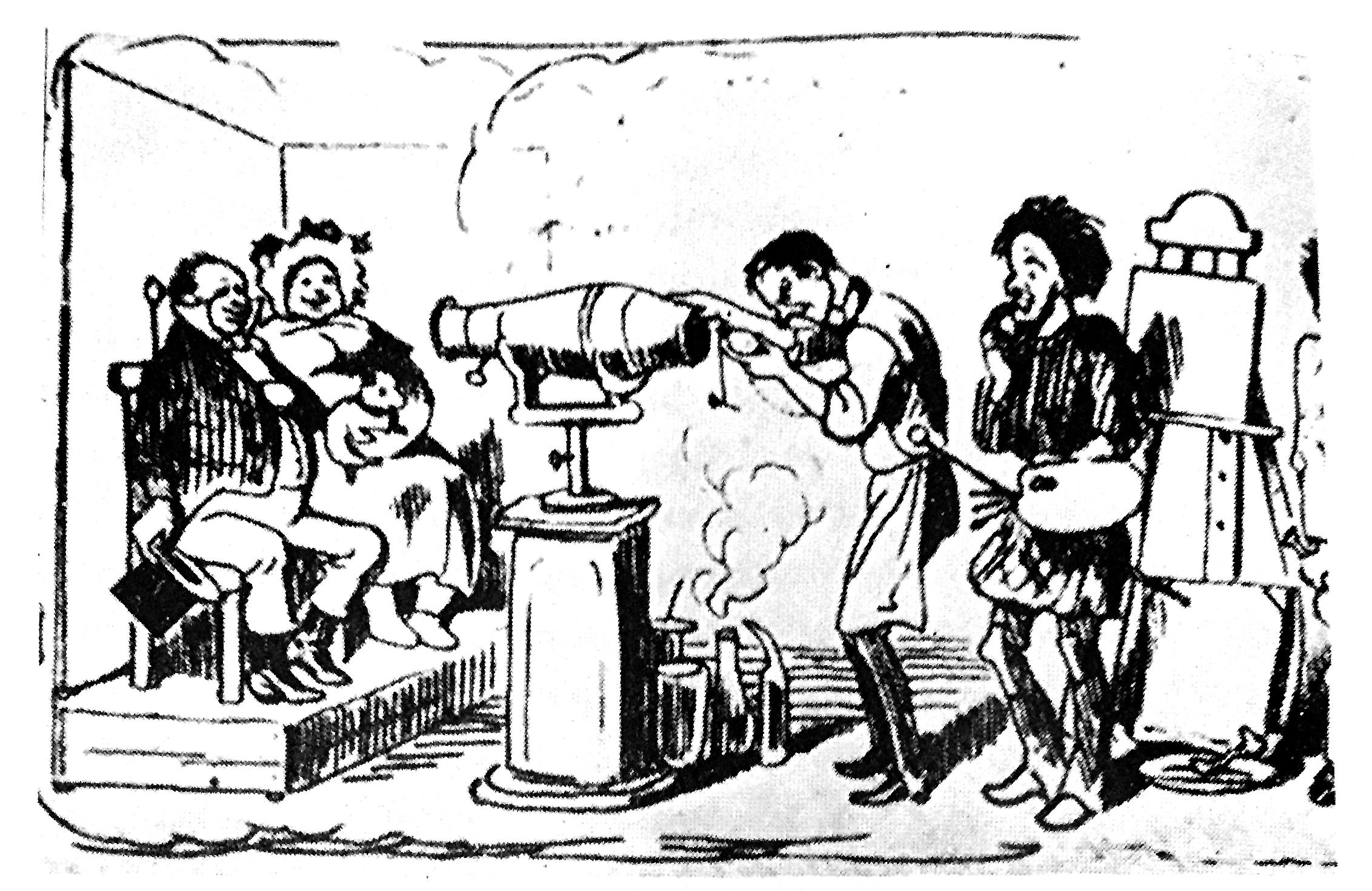}
    \caption{\textit{The Unhappy Painter}, by Theodor Hosemann, 1843, satirizes the painter, a victim of progress, made obsolete by a daguerreotype.} 
    \label{fig:hosemann}
\end{figure}

Another early use for the daguerreotype was to produce souvenirs for tourists: by 1850, daguerreotypes of Roman ruins completely replaced the etchings and lithographs  that tourists had previously purchased. As the technology improved, photography became indispensable as a source of records for engineering projects and disappearing architectural ruins, as well as for documentary purposes, such as Matthew Brady's photographs of the horrors of the American civil war.

``Is photography art?'' This question was debated for many decades, coalescing into three main positions. Many people believed that photography could not be art, because it was made by a mechanical device rather than by human creativity.
Many artists were dismissive of photography, and saw it as a threat to ``real art.'' For example, the poet Charles Baudelaire wrote, in a review of the Salon of 1859: ``If photography is allowed to supplement art in some of its functions, it will soon supplant or corrupt it altogether, thanks to the stupidity of the multitude which is its natural ally.''
A second view was that photography could be useful to real artists, such as for reference, but should not be considered as equal to drawing and painting. Finally, a third group, relating photography to established forms like etching and lithography, felt that photography could eventually be as significant an art form as painting.

Photography ultimately had a profound and unexpected effect on painting. 
Painters' mimetic abilities had been improving over the centuries.
Many painters of the 19th century, such as the Pre-Raphaelites like John Everett Millais and Neoclassicists like Ingres, painted depictions of the world with dazzling realism, more than had ever been seen before. However,  cameras became cheaper, lighter, and easier to use, and grew widespread among both amateurs and professionals. Realistic photographs became commonplace by the end of the 19th century. If photorealism could be reduced to a mechanical process, then what is the artist's role? 

This question drove painters away from visual realism toward different forms of abstraction.
James McNeill Whistler's Tonalist movement created atmospheric, moody scenes; he wrote: ``The imitator is a poor kind of creature. If the man who paints only the tree, or the flower, or other surface he sees before him were an artist, the king of artists would be the photographer. It is for the artist to do something beyond this.'' 
The Impressionists, who sought to capture the perceptions of scenes, were likely influenced by the ``imperfections'' of early photographs. In contrast,
Symbolists and post-Impressionist artists moved away from perceptual realism altogether.
Edvard Munch wrote ``I have no fear of photography as long as it cannot be used in heaven and in hell. ... I am going to paint people who breathe, feel, love, and suffer.'' 
Vincent Van Gogh, describing his artistic breakthroughs around 1888, wrote to his brother: ``You must boldly exaggerate the effects of either harmony or discord which colors produce. It is the same thing in drawing --- accurate drawing, accurate color, is perhaps not the essential thing to aim at, because the reflection of reality in a mirror, if it could be caught, color and all, would not be a picture at all, no more than a photograph.''  Photography continued to influence modern art of the 20th century; one can infer a significant influence of \'{E}tienne-Jules Marey's multiple-exposure photography on Futurism and Cubism, e.g., in Duchamp's \textit{Nude Descending A Staircase}.

\textit{It seems likely, in fact, that photography was one of the major catalysts of the Modern Art movement: its influence led to decades of vitality in the world of painting}, as artists were both inspired by photographic images and pushed beyond realism.

\begin{figure}
\centering
\includegraphics[width=3in]{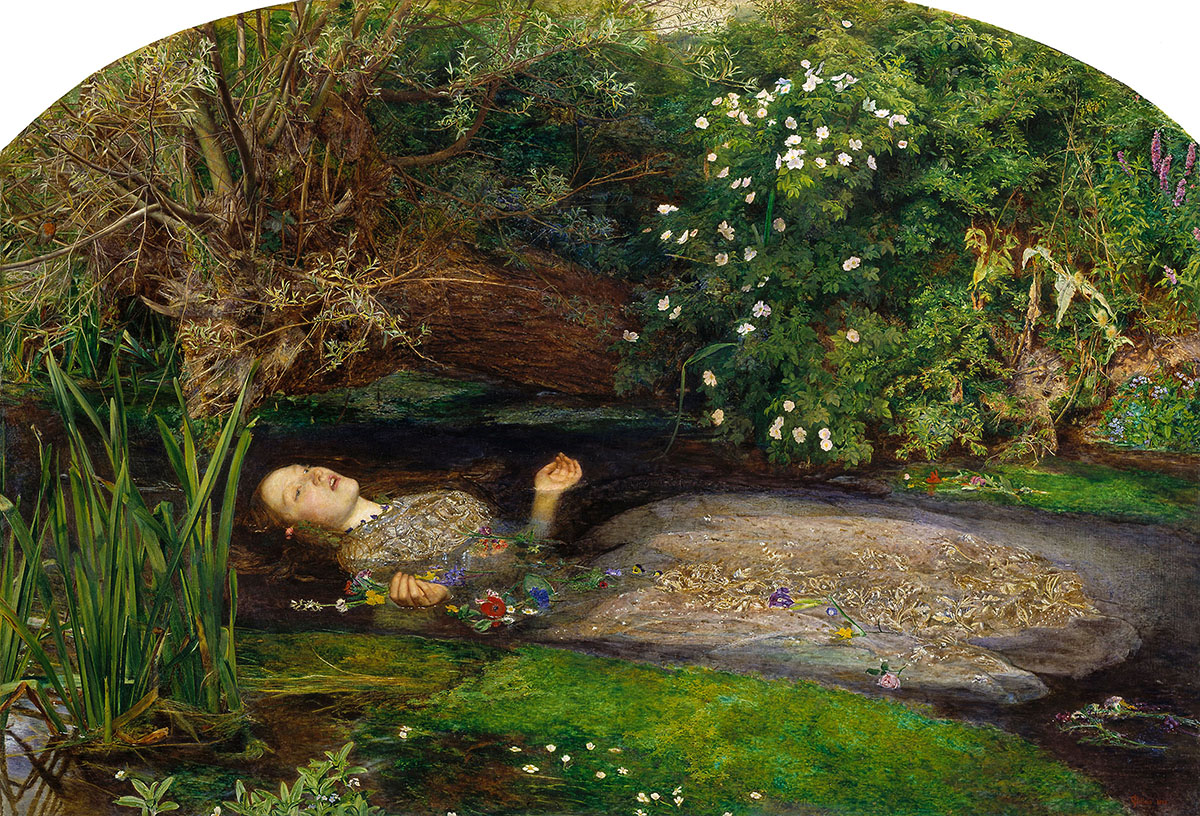}
\includegraphics[width=3in]{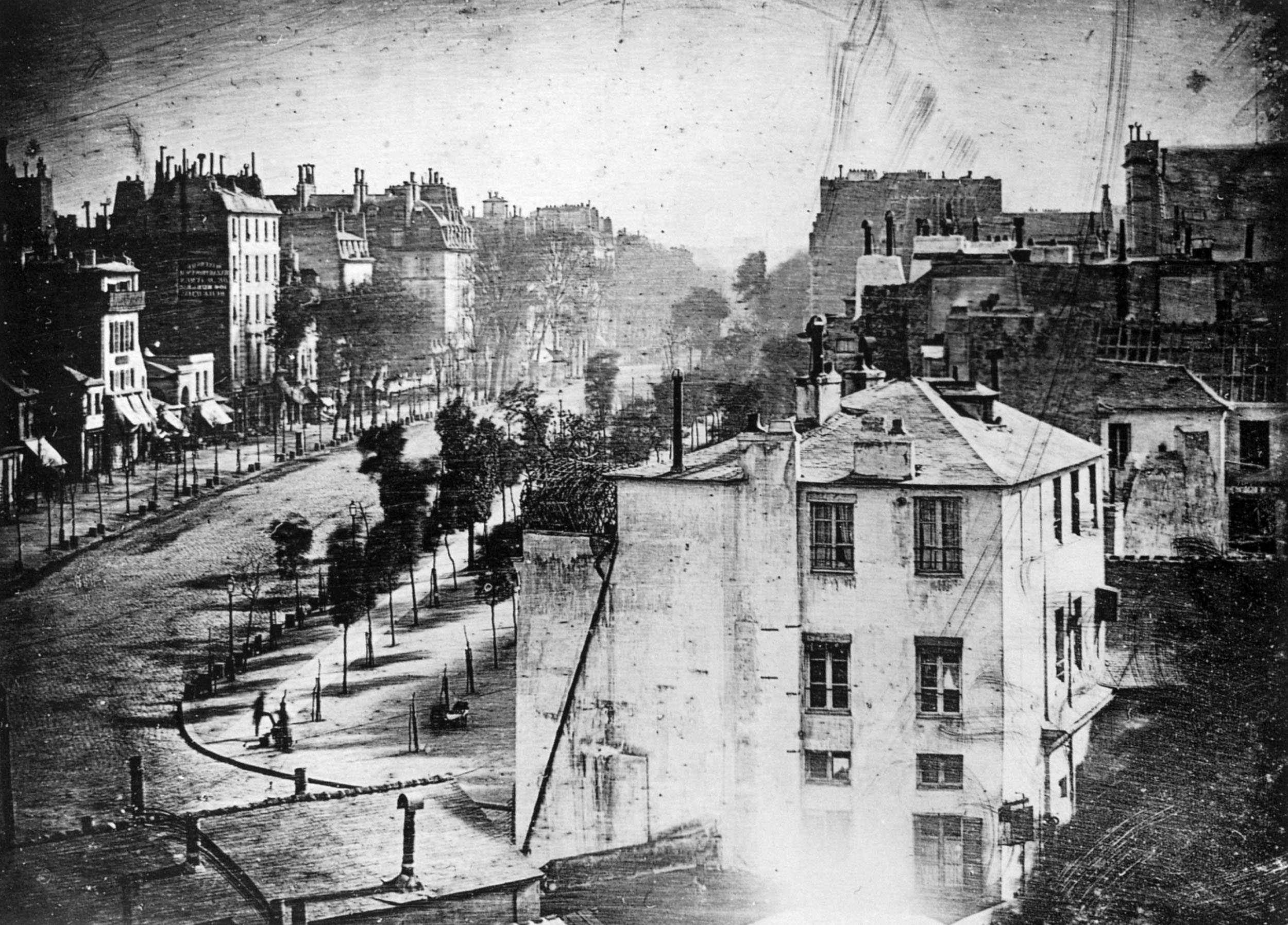} \\
\includegraphics[width=2in]{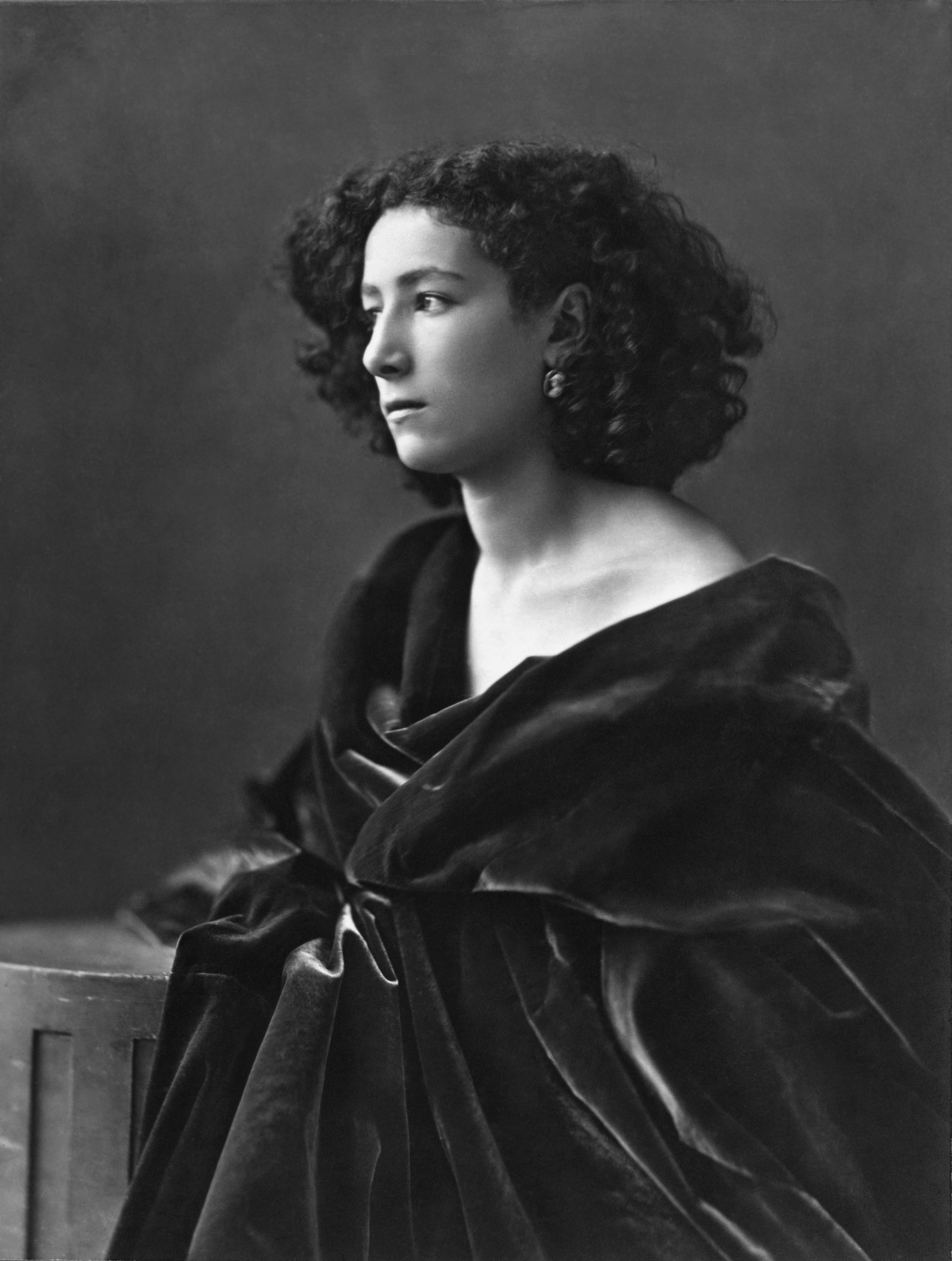}
\includegraphics[width=2in]{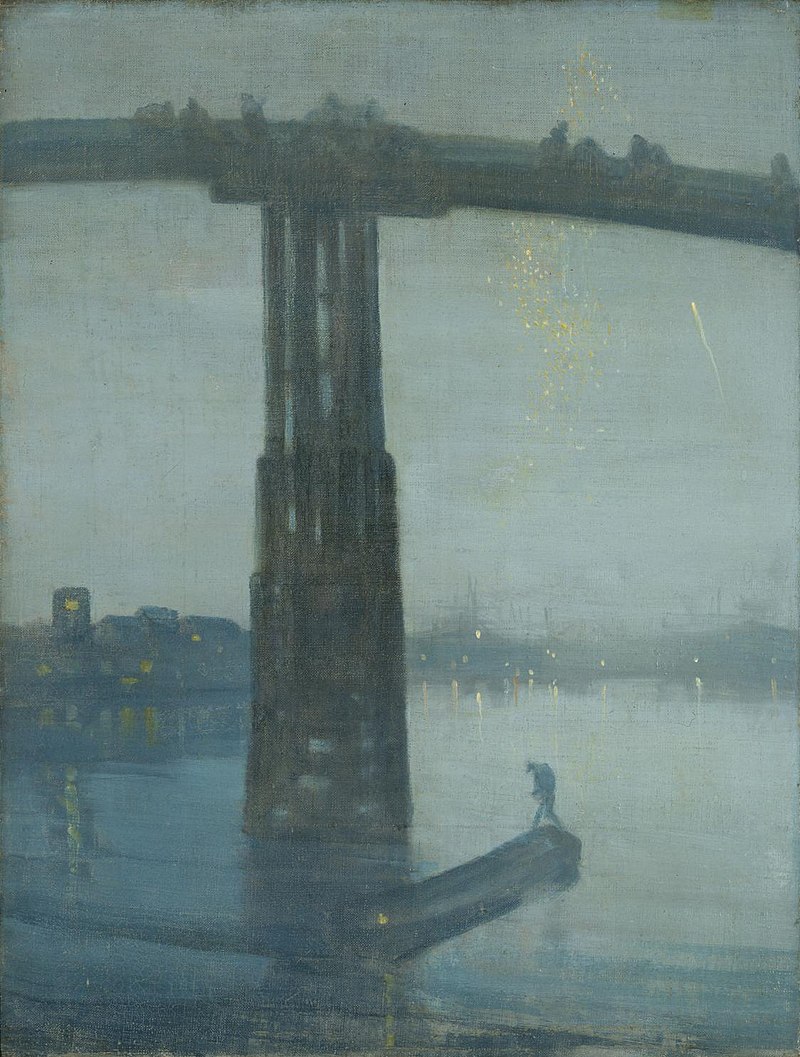}
\includegraphics[width=2in]{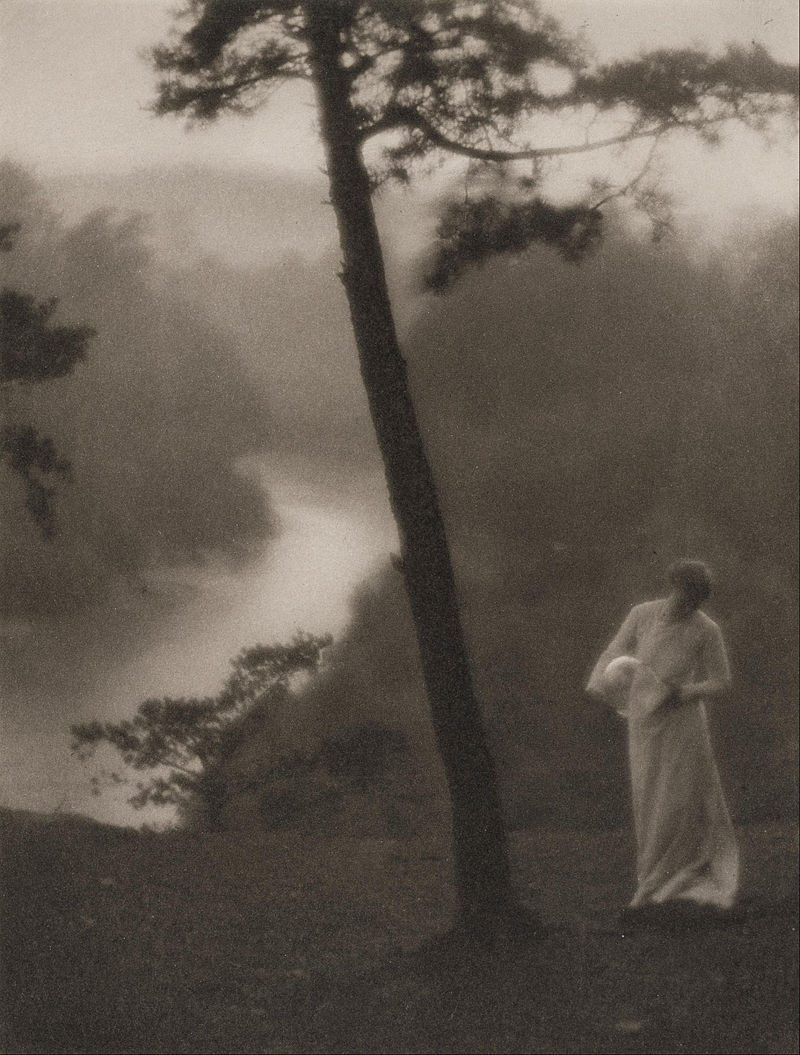}
\caption{
\label{fig:ptgphoto}
The interplay of painting and early photography.
(a) By the 19th century, Western painters had achieved dazzling levels of realism. (b) Early cameras took low-quality (though evocative) pictures. This daguerreotype took over ten minutes to expose. (c) However, camera technology steadily improved, capturing greater and greater realism, in much faster exposures. (d) This challenged painters to create works that were \textit{not} about hyper-realistic depiction, such as Whistler's Tonalist \textit{Nocturne}. (e) The Pictorialist photographers attempted to establish photography as an art form by mimicking the styles and abstraction of painting.
\textbf{Works:}
(a) \textit{Ophelia}, John Everett Millais, 1851
(b) \textit{Boulevard du Temple}, Daguerre, 1838;
(c) \textit{Portrait of Sarah Bernhardt}, F\'{e}lix Tournachon (Nadar), 1864;
(d) \textit{Nocturne in Blue and Gold: Old Battersea Bridge}, James MacNeill Whistler, c. 1872-1875;
(e) \textit{Morning}, Clarence H. White, 1908.
}
\end{figure}

Meanwhile, the Pictorialist movement, begun around 1885, was an attempt to firmly establish photography as an art form. Pictorialists introduced much more artistic control over the photographs, often using highly-posed subjects as in classical painting, and manipulating their images in the darkroom. Many of their works had a hazy, atmospheric look, similar to Tonalism, that softened the realism of high-quality photography. They seemed to be deliberately mimicking the qualities of the fine art painting of the time, and today much of their work seems rather affected.
They pursued various strategies toward legitimization of their work as art form, such as the organization of photographic societies, periodicals, and juried photography exhibitions \citep{Sternberger}.
Their works and achievements made it harder and harder to deny the artistic contributions of photography; culminating in the
``Buffalo Show,'' organized by Alfred Stieglitz at the Albright Gallery in Buffalo, NY, the first photography exhibition at an American art museum, in 1910. 
Photography was firmly established as an art, and free to move beyond the pretensions of Pictorialism.

This story provides several lessons that are directly relevant for AI as an artistic tool. At first, photography, like AI, was seen by many as non-artistic, because it was a mechanical process. Some saw photography as a threat and argued against its legitimacy. Photography did displace old technologies that had fulfilled non-artistic functions, such as portraiture's social function. Some artists enthusiastically embraced the new technology, and began to explore its potential. As the technology improved, and became more widespread over nearly a century, artists learned to better control and express themselves with the new technology, until there was no more real controversy over the status of photography. The new technology made image-making much more accessible to non-experts and hobbyists; today, everyone can experiment with photography. Furthermore, the new technology breathed new life into the old art form, provoking it toward greater abstraction. Wherever there is controversy in AI as an artistic tool, I predict the same trajectory. Eventually, new AI tools will be fully recognized as artists' tools; AI tools may stimulate traditional media as well, e.g., the New Aesthetic \citep{Sterling}.

\subsection{The technology of live-action cinema}

The story of filmmaking and technology has important lessons about how artists and technologists can work together, each pushing the other further. Most of the early photographers were, by necessity, both artists and technologists, experimenting with new techniques driven by their art or to inspire their art. But, in film and animation, this interaction has been much more central to the art form.

The history of film is filled with artist-tinkerers, as well as teams of artists and technologists. 
The Lumi\`{e}re Brothers created one of the first films, a simple recording of workers leaving their factory, but also experimented with a wide range of camera technologies, color processing, and artistic ways to use them. The stage magician George M\'{e}li\`{e}s filmed fantastical stories like \textit{A Trip to the Moon}, employing a wide range of clever in-camera tricks to create delightfully inventive and beguiling films. Walt Disney employed and pushed new technologies of sound and color recording, and drove other innovations along the way, such as the multiplane camera. Many of Orson Welles' innovative film techniques were made possible by new camera lenses employed by his cinematographer Gregg Toland. The introduction of more portable camera and audio equipment enabled the experiments of the French New Wave, who, in turn, influenced young American directors like Francis Ford Coppola and George Lucas.
George Lucas' team for Star Wars was an early developer of many new visual effects on a shoestring budget (think of Ben Burtt hitting telephone guy-wires to create the ``blaster'' sound effect), as well as an early innovation in digital film editing and compositing \citep{Droidmaker}. Digital and computer graphics technology, have, obviously, revolutionized film storytelling since then, with directors like Michel Gondry and James Cameron pushing the technology further into unforeseen directions. In each case, we see technologies rapidly adopted by directors to create new storytelling techniques and styles, transforming the medium over and over.


\newcommand{\theight}{1.2in}
\begin{figure}
    \centering
    (a) \includegraphics[height=\theight]{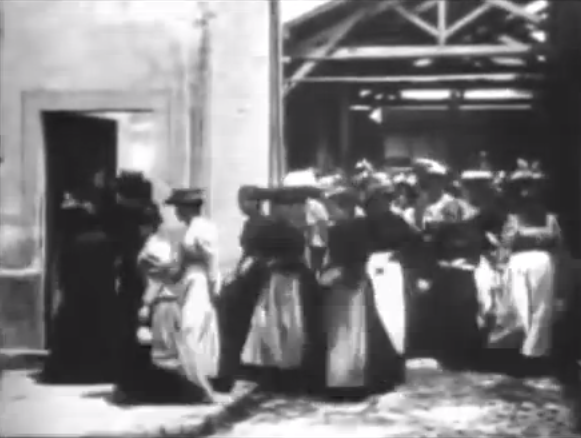}
 (b)    \includegraphics[height=\theight]{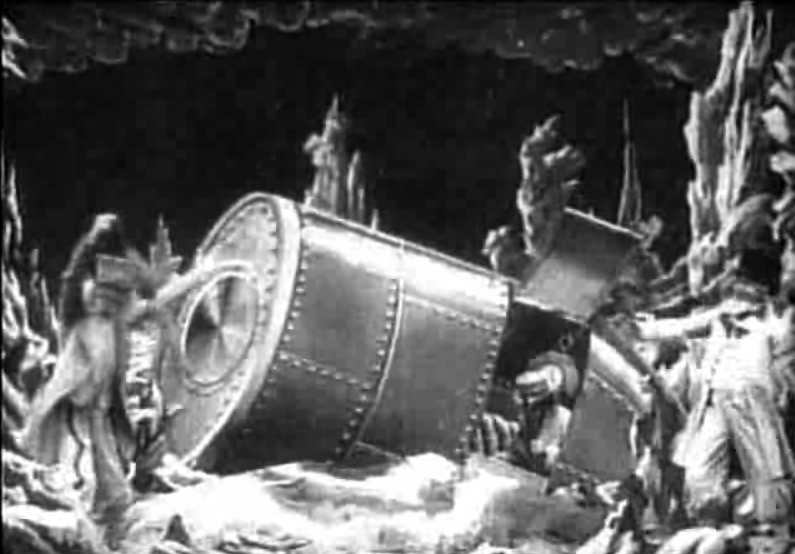}
(c)    \includegraphics[height=\theight]{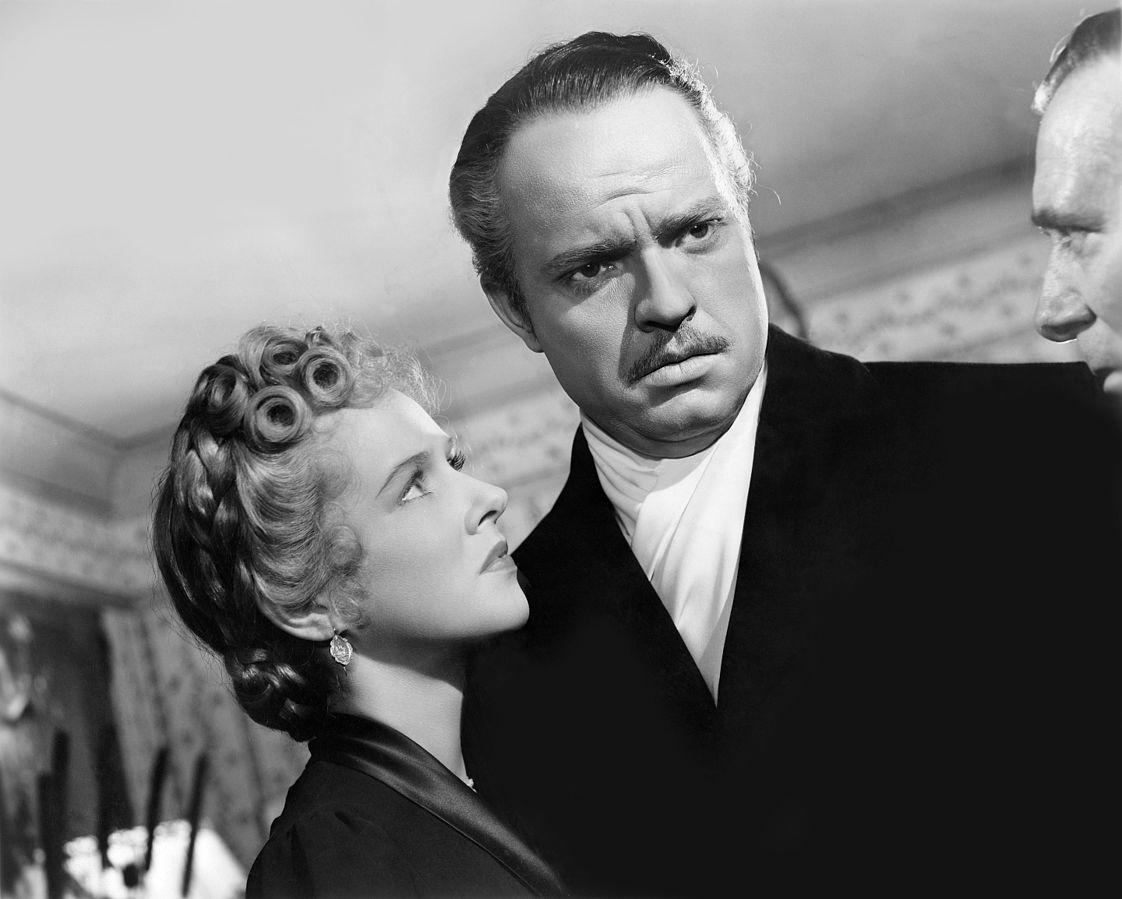}
    \caption{Technological developments in the art of filmmaking: (a) the first captured film, of workers leaving the Lumiere Brothers' factory (1895), (b)
    George M\'{e}li\`{e}s' 1902 \textit{A Trip to The Moon} filmed like a stage play but with fantastical special effects, (c) Citizen Kane, which used numerous experimental camera and lens effects to tell the story.
    \label{fig:movietech}
}
\end{figure}

\subsection{3D computer animation: a  collaboration}

3D computer animation as an artform was pioneered by
Pixar Animation Studios, and that success is due to the close collaboration of artists and engineers \citep{PixarTouch}. It all began with Ed Catmull, an animation enthusiast who received a PhD in computer science in 1974. In his thesis, he invented several core techniques that every major 3D computer graphics system uses today. During his time in graduate school, he quietly set a goal for himself: to make the world's first computer-animated film \citep{CreativityInc}. Consequently, he founded the Graphics Group at Lucasfilm Computer Division, and hired a team of brilliant engineers to invent computer systems to be used for film-making. However, none of this group could animate, that is, bring a character to life through movement. Hence, they recruited John Lasseter, an animator trained deeply in the Disney tradition. Through tight collaboration between Lasseter and the technical staff, they were able to invent new technologies and discover together how computer animation could start to become its own art form \citep{Lasseter}. This group, led by Catmull and Alvy Ray Smith, spun out as Pixar, and, over the following years, invented numerous technical innovations aimed at answering the needs set out by Pixar's  artists; in turn, the artists were inspired by these new tools, and pushed them to new extremes. One of their mantras was ``Art challenges technology, technology inspires art'' \citep{CreativityInc}.

Pixar, by design, treats artists and engineers both as crucial to the company's success and minimizes any barriers between the groups. When I worked there  during a sabbatical, despite my technical role, I had many energizing conversations with different kinds of artists, attended many lectures on art and storytelling, sketched at an open life drawing session, watched a performance of an employee improv troupe, and participated in many other social and educational events that deliberately mixed people from different parts of the company. 
This is the culture that, though it still has some flaws to address, achieved so many years of technical and creative innovation, and, ultimately, commercial and artistic success. 

Computer animation is another technology that scared traditional artists.
In the early days before they found Lasseter, the Lucasfilm Graphics Group made many attempts to interest Disney animators in their work \citep{PixarTouch}.
Smith later said: ``Animators were frightened of the computer. They felt that it was going to take their jobs away. We spent a lot of time telling people, ‘No, it's just a tool—-it doesn't do the creativity!' That misconception was everywhere'' \citep{StoryOfPixar}.
It is a common misconception that computer animation just amounts to the computer solving everything; a programmer presses a button and the characters just move on their own.  
In reality, computer animation is extraordinarily labor-intensive, requiring the skills of talented artists (especially animators) for almost every little detail. Character animation is an art form of extreme skill and talent, requiring laborious effort using the same fundamental skills of performance --- of bringing a character to life through pure movement --- as in conventional animation \citep{Lasseter}. 

Traditional cel animation jobs did not last at Disney, for various reasons. Disney Feature Animation underwent a renaissance in the early '90s, starting with \textit{The Little Mermaid}. Then, following some changes in management, the Disney animation began a slow, sad decline. After releasing duds like \textit{Brother Bear} and \textit{Home on the Range,} management shut down all traditional 2D animation at Disney, and converted the studios entirely to 3D computer animation.  Many conventional animators were retrained in 3D animation, but Disney's first 3D animation, \textit{Chicken Little} was still a dud.
Following Disney's acquisition of Pixar several years later, they  revived Disney's beloved 2D animation productions. The result, a charming and enjoyable film called \textit{The Princess and The Frog}, performed so-so at the box office, and, moreover, the animators' creative energy was focused on the newer 3D art form \citep{CreativityInc}. Today, traditional 2D animation at Disney is dead.\footnote{Traditional animation styles are still vital in countries like Japan and France that, unlike America, do not believe that animation is “just for kids.” Even so, their visual styles have evolved considerably due to computer technology.}
Today, computer animation is a thriving industry, and it thrives in many more places than cel animation ever did: at many different film studios, in visual effects for live-action films, in video games, television studios, web startups, independent web studios, and many more. There are now more types of opportunities for animators than ever before.
The story here is not the destruction of jobs, but the evolution and growth of an art form through technology.
This is another story that contradicts the popular notion of art and technology operating in conflict, when, in fact, the opposite is usually true. 

\subsection{Procedural artwork}
\label{sec:procedural}

\begin{figure}
\centering
(a)
\includegraphics[width=2in]{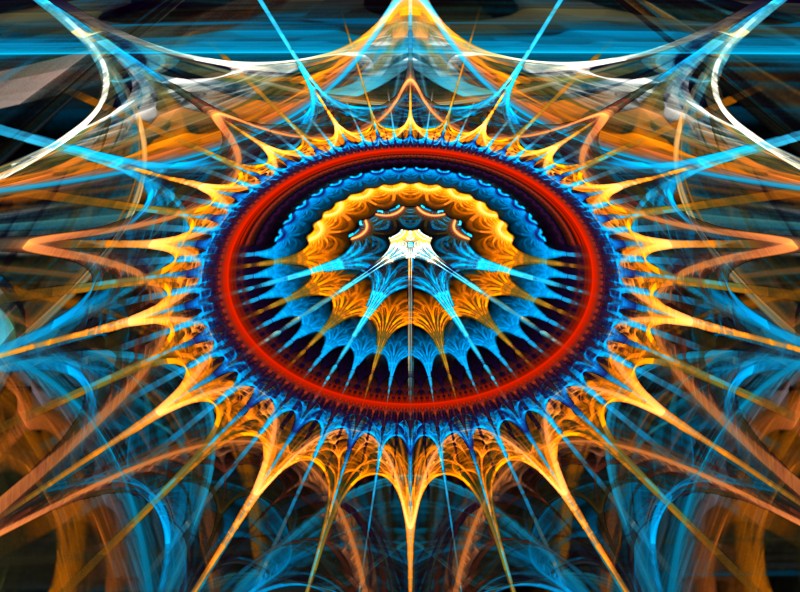}~
(b)
\includegraphics[height=1.48in]{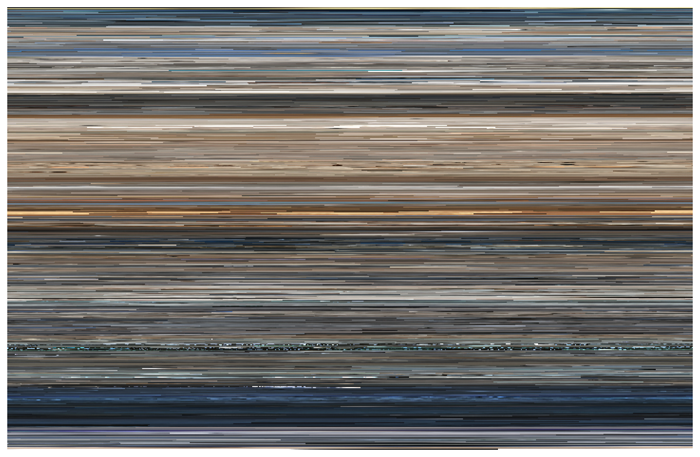}
\caption{\label{fig:procedural}
Procedural artworks in the fine art world of galleries and art museums.
(a) \textit{Electric Sheep}, by Scott Draves, evolves dazzling procedural abstract animations based on thousands of votes. 
``sheep'' by BrothaLewis.
(b) \textit{The Top Grossing Film of All Time, 1 x 1} (2000) by Jason Salavon, shows the average color of each frame of the movie Titanic. 
}
\end{figure}

In the art world, there is a long tradition of procedural artwork. Jean Arp created artworks governed by laws of chance in the 1910s (or so he claimed), and, beginning in the 1950s, John Cage used random rules to compose music. The term ``Generative Art'' appears to have originated in the 1960s. Sol LeWitt's wall drawings are provided as lists of precise instructions; people are still drawing new versions of his paintings after his death \citep{LeWitt}.
Starting in the 1970's, classically-trained painter Harold Cohen began exhibiting paintings generated by a program he wrote called AARON \citep{HaroldCohenAARON}. Since the 1980s, many current artists, such as Karl Sims, Scott Snibbe, Golan Levin, Scott Draves, and Jason Salavon, create abstract artworks by writing computer programs that generate either static images, or create interactive artistic experiences and installation works (Figure \ref{fig:procedural}). In Sims' and Draves' work, the artwork ``evolves'' according to audience input. The popularity of the Processing computer language for artists speaks to the growth of procedural art.

In each of these cases, despite the presence of procedural, emergent, and/or crowdsourced elements, the human behind it  is credited as the author of the artwork, and it would seem perverse to suggest otherwise.  The human has done all of the creative decision-making around the visual style, of designing a framework and process, of testing and evaluating alternative algorithms, and so on.

\subsection{State of the art in computer science research}
\label{sec:NPR}

Recent developments in computational artistic image synthesis are quite spectacular. But they should not be mistaken for AI artists. 

\begin{figure}
\centering
(a) \includegraphics[width=4in]{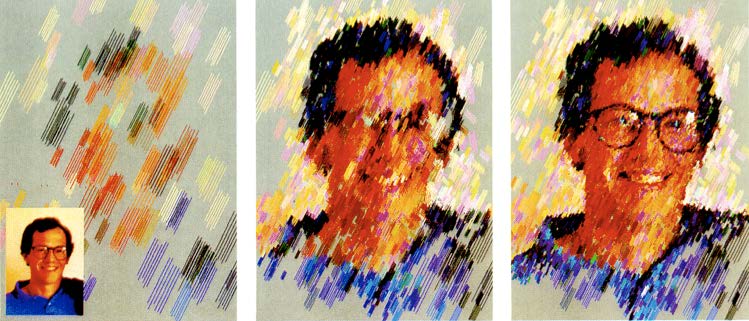} \\[2ex]
(b) \includegraphics[width=2in]{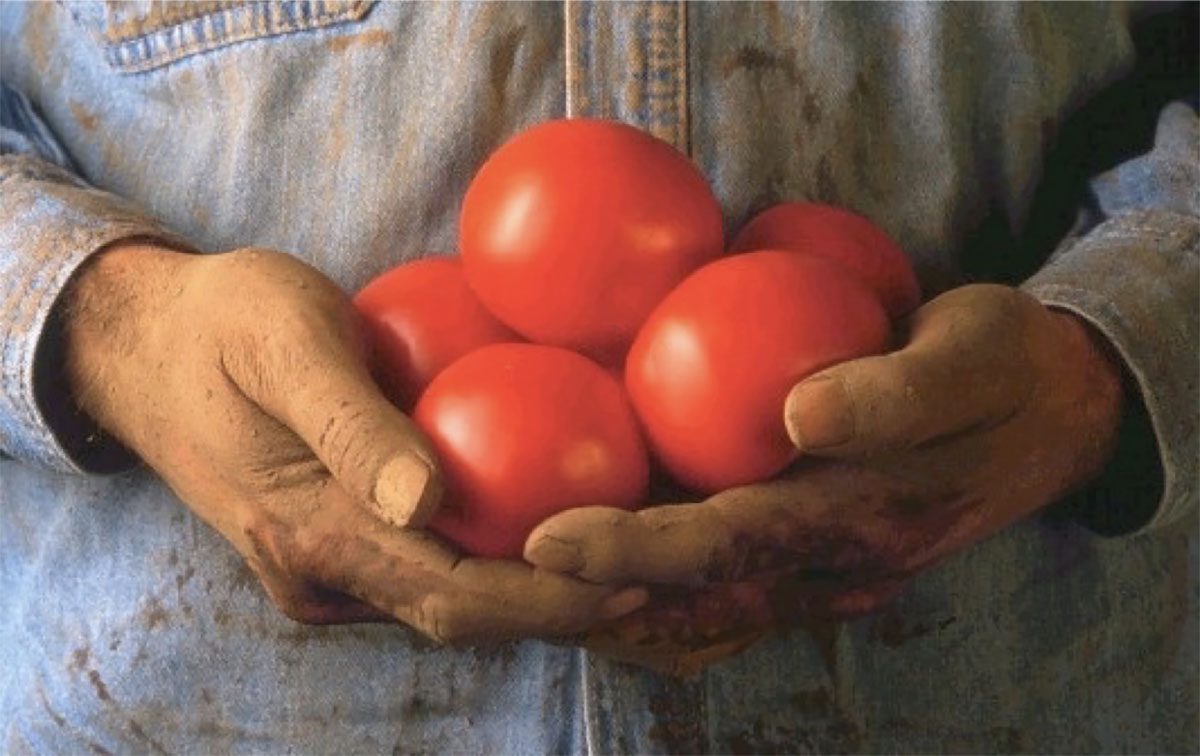}
\includegraphics[width=2in]{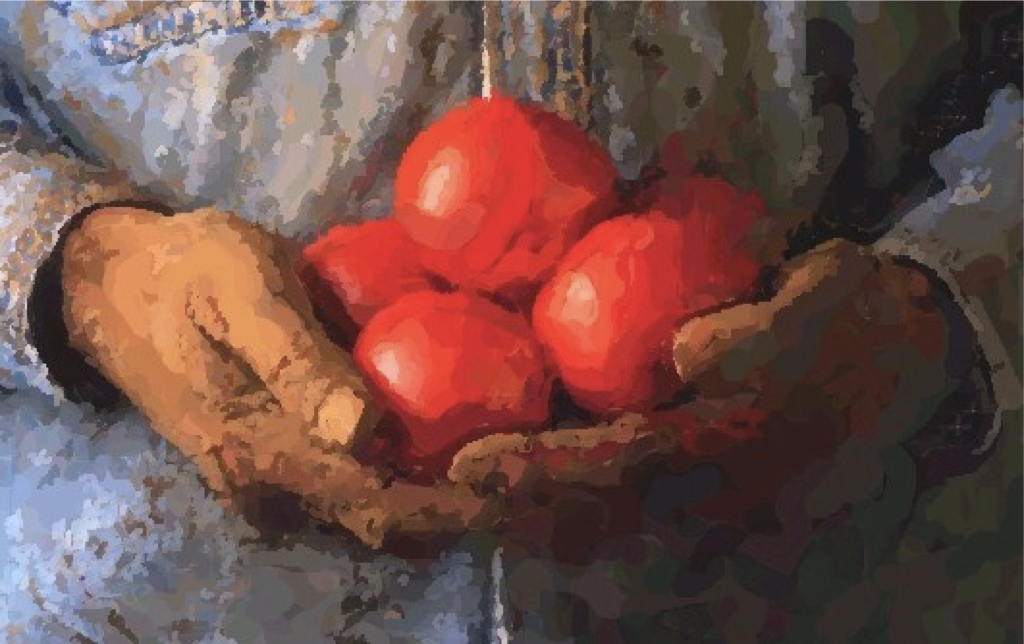}
\caption{\label{fig:painterly}
Painterly rendering algorithms that process an input photograph, using hand-coded rules and algorithms.
(a) Paul Haeberli's interactive painting system \citep{Haeberli}.
(b) My automatic painting system 
\citep{Hertzmann1998} processes a photograph without user input aside from selecting some parameter settings.
}
\end{figure}

Non-Photorealistic Rendering (NPR) is a subfield of computer graphics research  \citep{NPRbook} that I have worked in for many years. NPR research develops new  algorithms and artistic tools for creating images inspired by the look of a conventional media, such as painting or drawing. Paul Haeberli's groundbreaking 1990 paper \citep{Haeberli} introduced a paint program that began with a user-selected photograph. Whenever the user clicked on the canvas (initially blank), the system placed a brush stroke with color and orientation based on the photograph. In this way, a user could quickly create a simple painting without any particular technical skill (Figure \ref{fig:painterly}(a)). In a follow-up paper, Pete Litwinowcz automated the process entirely, by placing brush strokes on a grid \citep{Litwinowicz}. My own first research paper arose from experimenting with modifications to his algorithm: the method that I came up with creates long, curved strokes, beginning with large strokes that were then refined by small details \citep{Hertzmann1998} (Figure \ref{fig:painterly}(b)). The algorithm was inspired by my experience with real painting, and the way artists often start from a rough sketch and then refine it.

This type of artistic algorithm design reflects the majority of computer graphics research in this area \citep{hertzmannSBR,NPRbook}. The algorithms are automated, but we can explain in complete detail why the algorithm works and the intuitions about artistic process it embodies. This mathematical modeling of artistic representation continues the investigations begun in the Renaissance with Filippo Brunelleschi's invention of linear perspective, a viewpoint I have written about elsewhere \citep{HertzmannScienceOfArt}.

\begin{figure}
    \centering
        \includegraphics[width=2in]{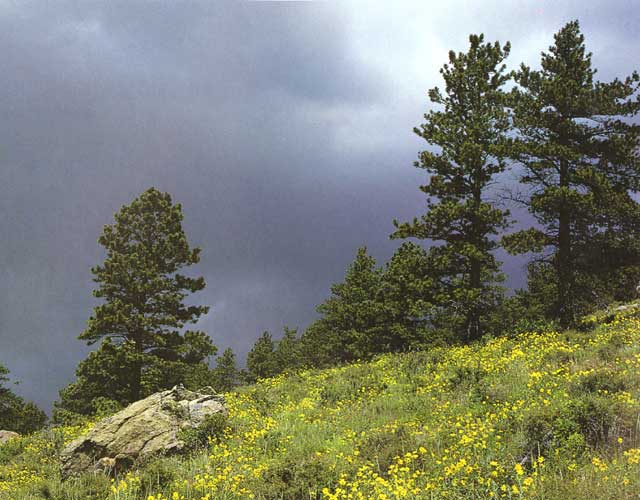}
        \includegraphics[width=2in]{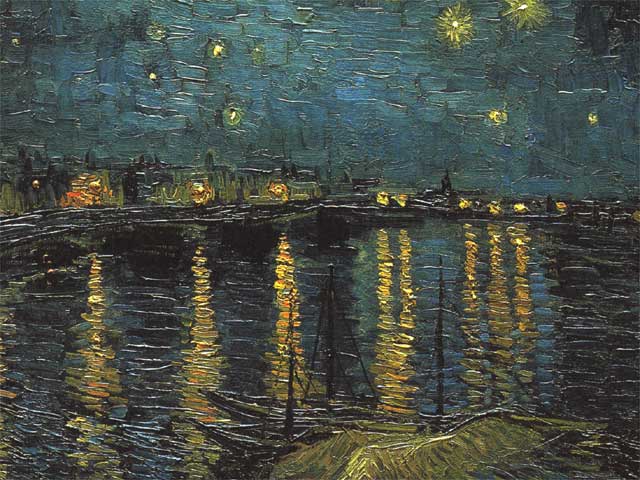}
        \includegraphics[width=2in]{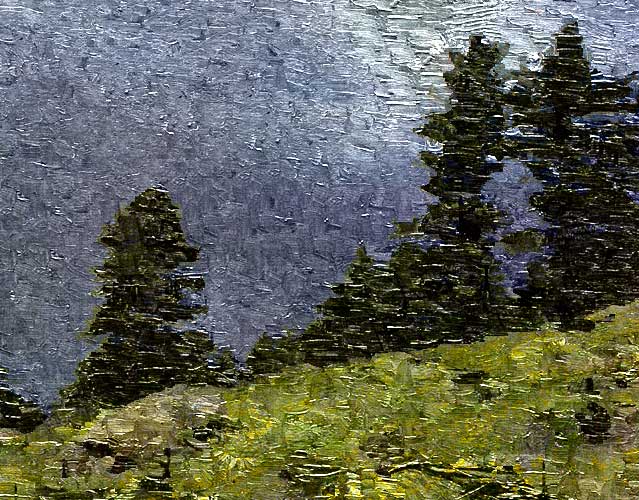}
    \caption{
\label{fig:analogies}
Our Image Analogies algorithm \citep{HertzmannImageAnalogies}, which stylizes a photograph in the style of a given artwork; in this case, Van Gogh's \textit{Starry Night over the Rh\^{o}ne}.  
    }
\end{figure}

At some point, I found it very difficult to embody richer intuitions about artistic process into source code. Instead, inspired by recent results in computer vision \citep{EfrosLeung}, I began to develop a method for working from examples. My collaborators and I  published this method in 2001, calling it ``Image Analogies'' \citep{HertzmannImageAnalogies}. We presented the work as learning artistic style from example. But the ``learning'' here was quite shallow. It amounted to rearranging the pixels of the source artwork in a clever way, but not generalizing to radically new scenes or style (Figure \ref{fig:analogies}). (A related style transfer method was published concurrently by Efros and Freeman \cite{EfrosFreeman}.) Since then, other researchers have improved the method substantially, making it much more robust \citep{StyLit}.

\begin{figure}
    \centering
    \includegraphics[width=2in]{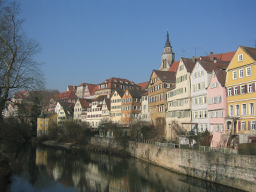}
    \includegraphics[width=2in]{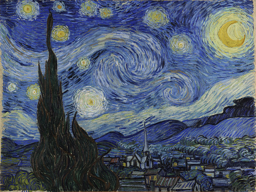}
    \includegraphics[width=2in]{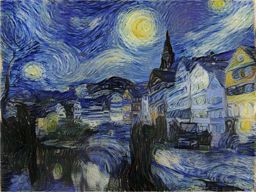}
\caption{The Neural Style Transfer Algorithm \citep{GatysNeuralStyle}, which stylizes a photograph in the style of a given artwork; in this case, Van Gogh's \textit{The Starry Night}. This algorithm has led to numerous new apps and research in stylization. 
    \label{fig:gatys}
    }
\end{figure}

In 2016, Leon Gatys and his colleagues published a new breakthrough in this space, Neural Style Transfer \citep{GatysNeuralStyle}. Based on recent advances in neural networks, their method transfers certain neural network correlation statistics from a painting to a photograph, thus producing a new painting of the input photograph (Figure \ref{fig:gatys}). The method is still ``shallow'' in a sense --- there is no ``understanding'' of the photograph or the artwork --- but the method seems to be more robust than the original Image Analogies algorithm. This paper led to a flurry of excitement and new applications, including the popular Prisma app and Facebook's Live Video stylization, as well as many new research papers improving upon these ideas. This work is ongoing today.

\begin{figure}
    \centering
        \includegraphics[width=5in]{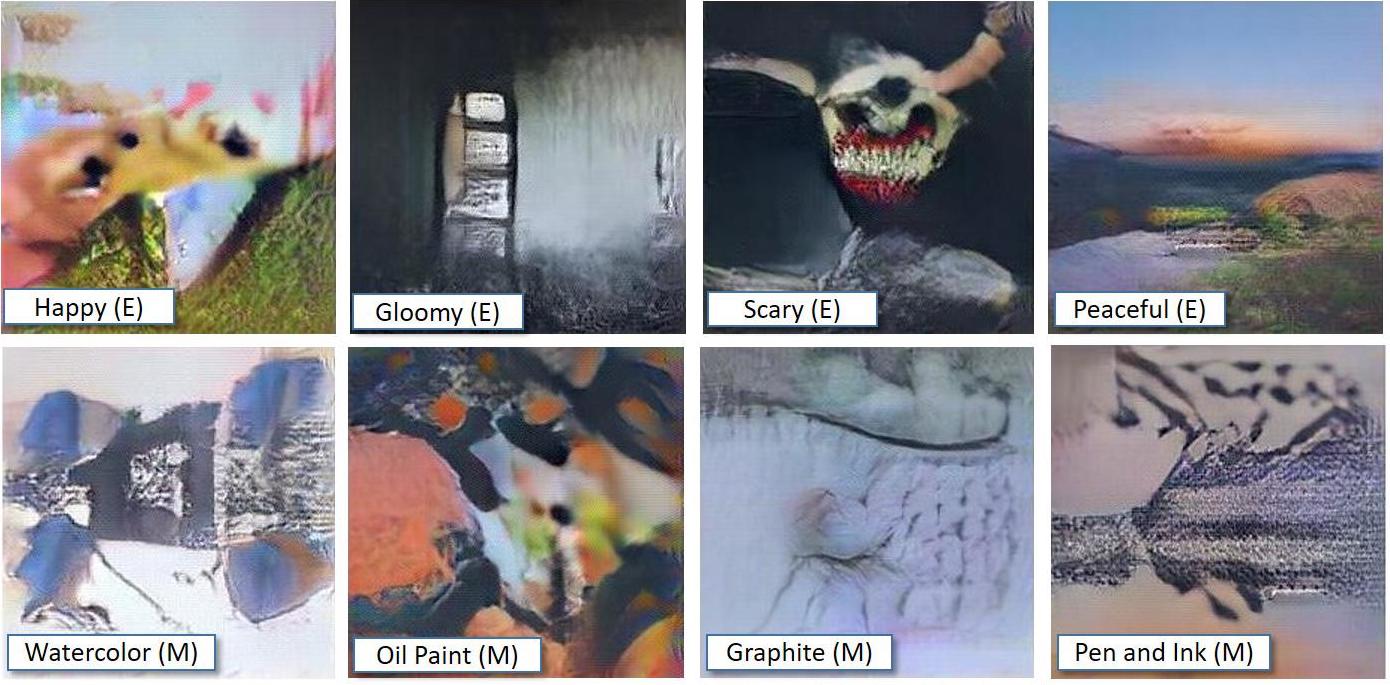}
    \caption{Images that we generated using a neural network trained on different subsets of a database of artistic imagery \citep{BAM}. Each one is meant to typify a single stylistic category or a medium, though the biases of the convolutional neural network architecture are also evident.
    \label{fig:bamgan}
    }
\end{figure}

Another development which received considerable attention in 2015 was the invention of DeepDreams by Mordvintsev et al.~\cite{DeepDreams}, who, developing a visualization tool for neural networks, discovered that a simple activation excitation procedure produced striking, hallucinatory imagery of a type we had never seen before.
There are many other current projects, particularly those around Generative Adversarial Networks \citep{GANs} and Project Magenta at Google \citep{NYTimes_Magenta}, that also show promise as new artistic tools. For example, Figure \ref{fig:bamgan} shows images that we generated by visualizing trends learned by a neural network from a large collection of artistic images in different styles. A variety of related images are produced by Creative Adversarial Networks \citep{CAN}; it is a visual style that seems familiar but not the same as what we are familiar with (and is probably driven, in part, by the biases of the convolutional neural network representation). 

\textit{In each of these cases, the artworks are produced by a human-defined procedure, and the human is the author of the imagery.}

\subsection{Artificial Intelligence is not intelligent}
\label{sec:ai}

Unfortunately, there has been a considerable amount of media hype around AI techniques. In the news media, algorithms are often anthropomorphized, as if they have the same consciousness as humans (e.g., \citep{fairbots}), and sometimes they are described  as artists \citep{MSartist,shimon,pogue}. 
In fact, we do not really know what consciousness is (despite many theories), or what it would mean to embody it in an algorithm. 

Today, the most-successful AI and machine learning algorithms are best thought of as glorified data-fitting procedures \citep{Brooks}. That is, these algorithms are basically like fitting a curve to a set of datapoints, except with very sophisticated ways to fit high-dimensional curves to millions of datapoints.
When we as researchers speak of ``training'' an algorithm, or an algorithm that ``learns,'' it is easy to misinterpret this as being the same thing as human learning. But these words mean quite different things in the two contexts.
In general, ``training'' a model to learn a task involves careful human effort to formulate the problem, acquire appropriate data, and test  different formulations. It is laborious and requires considerable expertise and experimentation. When a new task needs to be solved, the human starts over.

Compared to human intelligence, these algorithms are brittle and bespoke. For example, image recognition algorithms have undergone breathtaking breakthroughs in the past decade, are now widely used in consumer products. Yet they often fail on inputs that suggest bizarre misunderstandings; the existence and robustness of adversarial examples \citep{intriguing} and procedurally-evolved images \citep{InnovationEngine} demonstrates that these algorithms have not really learned anything like human-level understanding.  
They are like tourists in a foreign country that can repeat and combine phrases from the phrasebook, but not truly understand the foreign language or culture. These systems have no autonomy except within the narrow scope for which they were trained, and typically fail-safes must be put in place as well, e.g., Google Photos no longer classifies anything as a ``gorilla'' because of one high-profile failure \citep{gorilla}.


There are some fascinating parallels between human learning and machine learning, and it does seem likely that humans are, in some way, optimized by evolutionary principles \citep{optimaForAnimals,KORDING2006319,howToGrowAMind,gopnik}. But going from these high-level analogies to actual machine intelligence is a problem for which the solution is not even on the horizon.


\section{How technology changes art}

Based on this history, I now make several specific claims about how technology changes art. Far from replacing artists, new technologies become new tools for artists, invigorating and changing art and culture.  These claims apply equally to current developments in AI as they did to previous developments like photography and animation.

\subsection{Algorithms are artists' tools}
\label{sec:artistTools}

\textit{``my watercolor teacher used to say: let the medium do it.  true that --- so my sketch provides the foundation and then the network does it thing; i don't fight, just constantly tweak the \#brushGAN toolkit'' -- Helena Sarin (@glagolista)} 

In every technology that we currently employ --- whether photography, film, or software algorithm --- the technologies and algorithms we use are basic tools, just like brushes and paint.\footnote{This view reflects, I think, the conventional wisdom in the computer graphics research field, which is my research background. This field has always had close ties with certain artistic communities, especially computer animation and visual effects, and, based on this experience, the field is often  resistant to attempts to automate creative tasks. In contrast, artificial intelligence researchers use terminology much more aspirationally, historically using words like ``intelligence,'' ``learning,'' and ``expert systems'' in ways that far simpler than the human versions of these things.}
The same is true for the new AI-based algorithms that are appearing. They are not always predictable, and the results are often surprising and delightful --- but the same could be said for the way watercolor flows on the page. There is no plausible sense in which current systems reflect ``true'' artificial intelligence: there is always a human behind the artwork.

\begin{figure}
    \centering
    \includegraphics[height=1.5in]{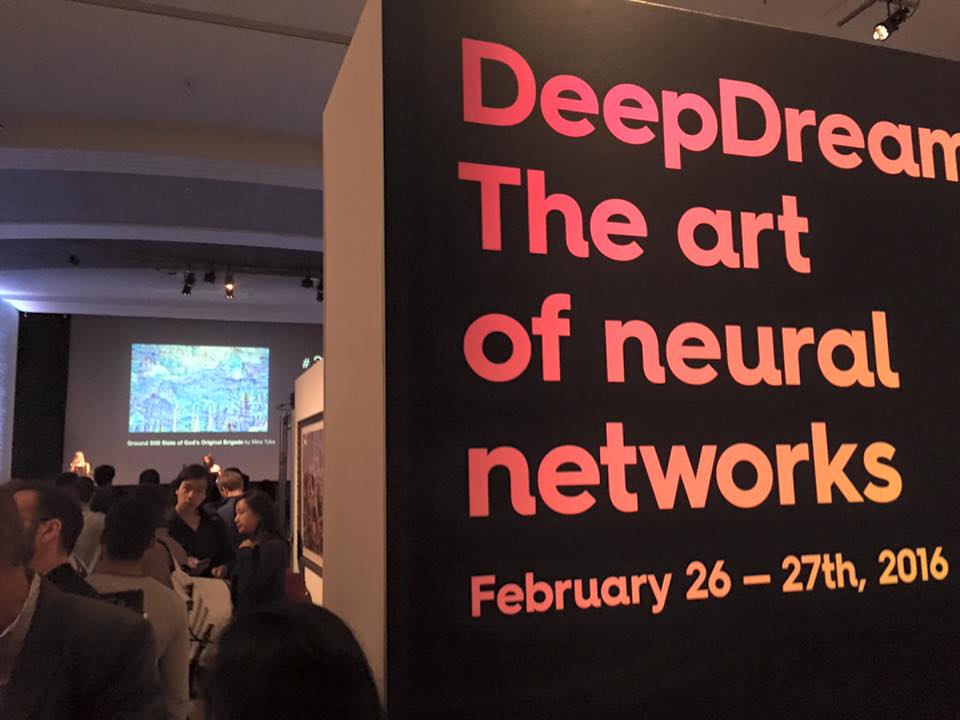}~~
    \includegraphics[height=1.5in]{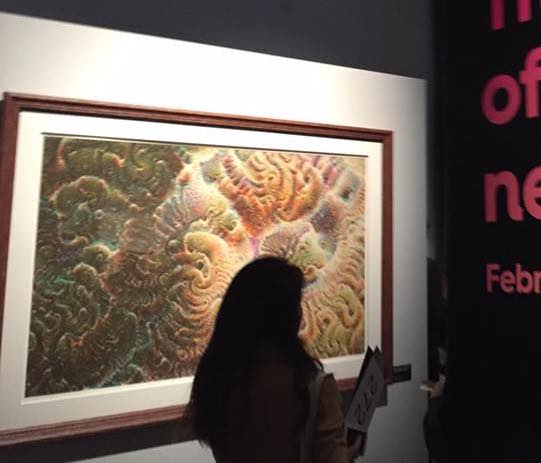}
    \caption{Neural network art exhibition at Gray Area in San Francisco}
    \label{fig:grayarea}
\end{figure}

Applying the same standard to the current research in neural networks and neural style transfer, it would seem equally perverse to assign authorship of their outputs to the software itself. The DeepDream software was authored by a human; another human then selected an input image, and experimented with many parameter settings, running the software over and over until obtaining good results.  Indeed, in a recent art exhibition meant to promote these methods and their exploration \citep{grayarttyka} (Figure \ref{fig:grayarea}), human artists were credited for each of the individual works.

The same process of selection of tools and inputs, adjusting settings and even modifying code, and iterating until a desirable output is produced, occurs in all current and forseeable computer artworks. Computer-generated artworks result from considerable time and effort from human artists, from conception of the idea, to painstakingly guiding the execution, to selecting from among the  outputs.

There have been a few cases where AI algorithms have been presented as artists or potential artists \citep{PaintingFool,InnovationEngineCACM,shimon,MSartist,pogue}. For the above reasons, I think this claim misunderstands the nature of procedural art.

In short, in our present understanding, \textit{all art algorithms, including methods based on machine learning, are tools for artists; they are not themselves artists}.

\subsection{New technology helps art stay vital}

Rather than being afraid of the new technologies, we should be enthusiastic about the new artworks that they will enable artists to produce. When we think of art of having external influences, we normally think of social or political influences, but ignore the effect of new tools.  
In contrast, I argue that, especially from 19th century onwards, technological developments have played a pivotal role in advancing art, in keeping it vital and injecting fresh ideas.  The stories I gave of photography and cinema include many  examples of this. However, the effect is far more widespread.  

\begin{figure}
    \centering
    (a)
    \includegraphics[width=2in]{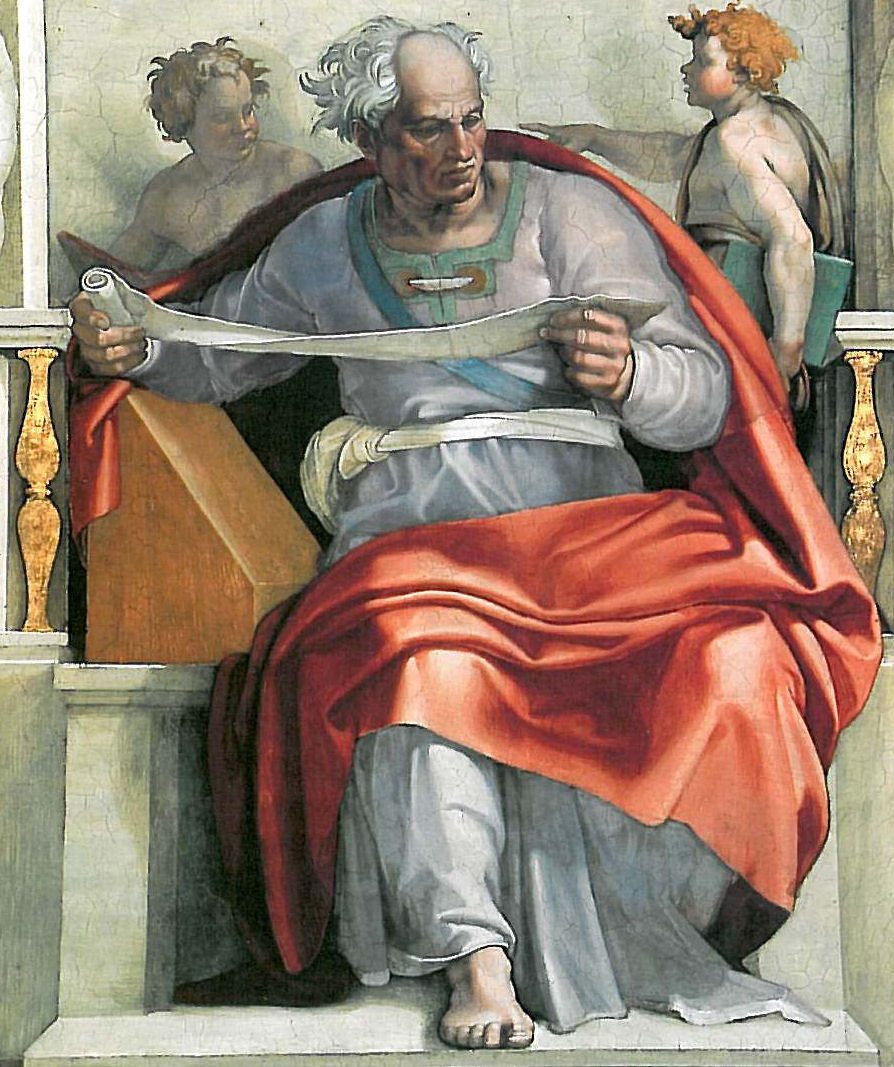}
    (b)
    \includegraphics[height=2.387in]{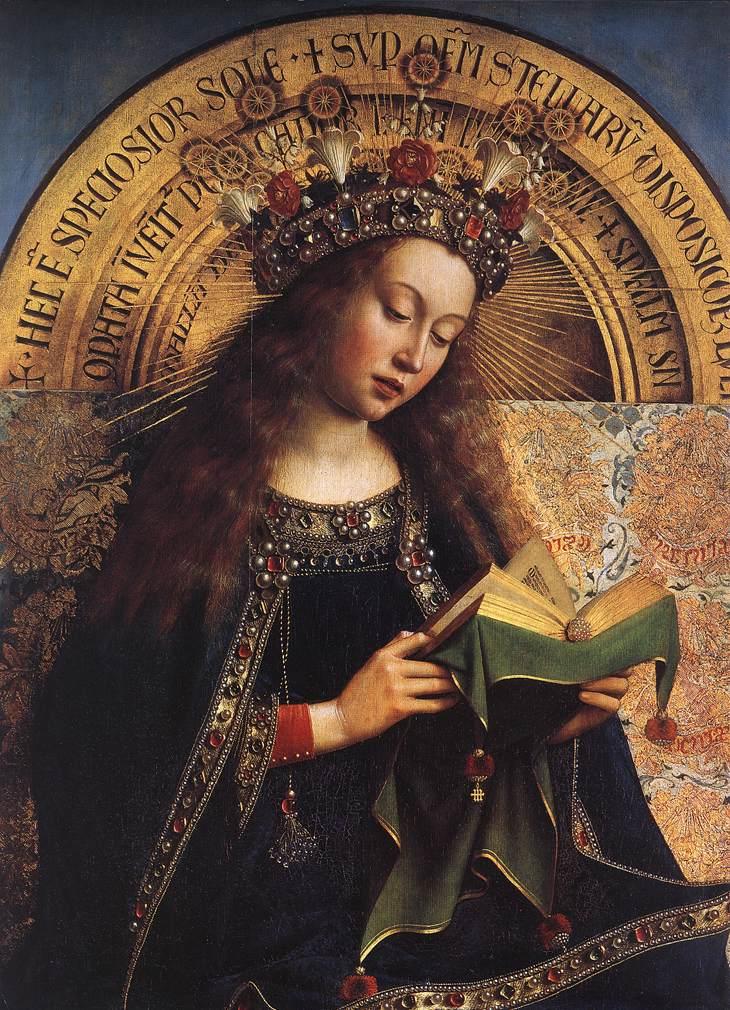}
    \caption{
    The development of the oil painting technology changed painting as an art form.  
    (a) Fresco painting by Michaelangelo on the Sistine Chapel (ca.~1508). The fresco process was difficult and achieved limited tonal range. Today, fresco is defunct as a medium.
    (b) Oil painting by Jan van Eyck for the Ghent Alterpiece (ca.~1430). Much richer colors and lighting are possible with oil paint.
    }
    \label{fig:eyck}
\end{figure}

One of the most important breakthroughs in the history of Western art was the invention of oil paint by Flemish painters such as Jan van Eyck in the 15th century \citep{Gardner}. Previously, painting had been done primarily with tempera, which lacks subtle coloration, and fresco, which was very cumbersome to work with. Oil paint had existed in some form for centuries, but van Eyck and others found new techniques that gave them a very practical new medium. It was fast-drying and allowed rich colors and tones, sharp edges, and hard surfaces. The rich light and color that we associate with the Northern Renaissance and the Italian Renaissance are due to this technology (Figure \ref{fig:eyck}).

In each decade since the 1950s, many of our culturally-important works used technology that had only been invented within the previous ten years. For example, most technology used in today's feature films did not exist ten years ago (e.g., widespread use of HD digital cameras; facial performance capture); the same goes for artworks using smartphones and crowdsourcing; artworks involving white LEDs and Arduino controllers; DJs performing on stage behind their laptops; and so on. Even the most v\'{e}rit\'{e}-seeming romantic comedies frequently involve recent digital video editing and digital backdrops. 

Conversely, artistic styles that fail to change become stale and lose their cultural relevance; the adoption and exploration of new technology is one of the ways that art stays vibrant. For example, the introduction of synthesizer music into 1980s pop music created a new sound that was exciting and modern. The sound diversified as the tools improved, until grunge became popular and made the 80s synthpop sound seem superficial and old-fashioned. Nowadays, a recent revival of 70s and 80s instruments by bands like Daft Punk and LCD Soundsystem seems most exciting at times when they are creating new types of music using old instruments. In contrast, the swing music revival of the 1990s never went anywhere (from bands like Big Bad Voodoo Daddy and Squirrel Nut Zippers) in my opinion, because the bands mimicked classic styles with classic instruments, without inventing anything particular original themselves.

In each era, radical technological innovations are met by artists with both enthusiasm and rejection. For example, when the Moog synthesizer became popular, it was adopted by big-name bands like Emerson, Lake, and Palmer. Other bands felt that twisting knobs to make music was “cheating:” Queen's album covers proudly state that the band did not use synthesizers. Robert Moog described one New York musician who said of the instrument ``This is the end of the world'' \citep{GuardianMoog}. 
It now seems silly to imagine that people might have ever categorically objected to synthesized music, or to the scratching and sampling of hip-hop DJs, just as it now seems silly that people once rejected waltzing, the Impressionists, and the Rite of Spring as invalid or immoral.

In addition to stimulating professional artists, new tools make art more accessible to larger portions of society. Photography was once accessible only to the most determined early adopters, but has continually become easier, faster, and more compact, to the point where nearly everyone carries a mobile phone camera in their pocket or purse. The same goes for the tools of cinematography (from hand-cranked to heavy cameras to Steadicams to handycams to iPhones), and so on. Modern computers give nearly everyone access to digital equivalents of darkrooms, mixing studios, painting studios, and so on; these were formerly highly-specialized technologies requiring laborious effort.

\begin{figure}
    \centering
    \includegraphics[width=1.25in]{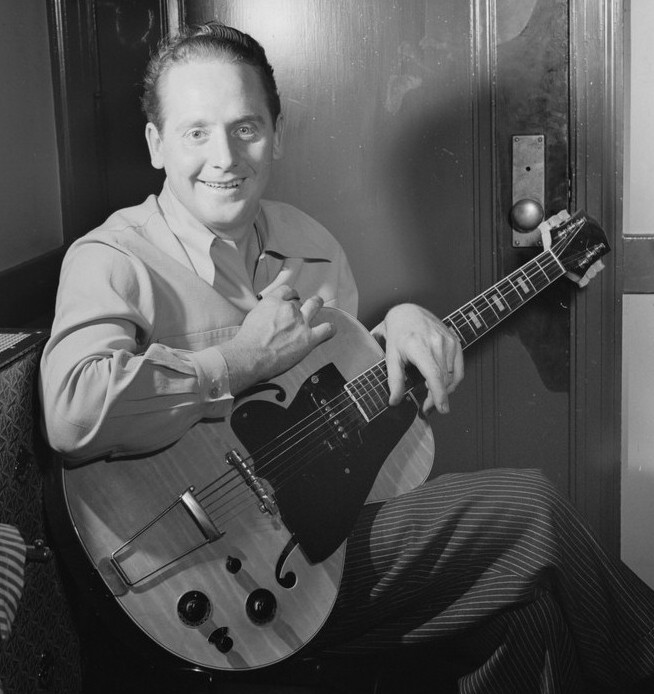}
    \includegraphics[width=2in]{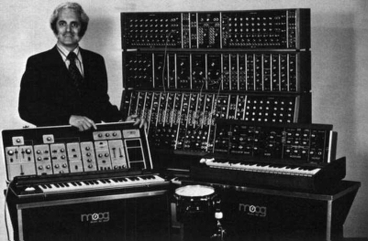}
    \caption{
    \label{fig:paul_moog}
    Les Paul, inventor the solid-body electric guitar in 1943; Robert Moog, pioneer of the electronic synthesizer, respectively. Their technologies transformed popular music in ways they could not have foreseen.
    }
\end{figure}

\subsection{New technology does not cause net unemployment}

Concerns of how technology displaces jobs have been around since at least the 19th century, when Luddite protesters destroyed mechanical weaving machines, and, in folk songs, John Henry competed against a steam drilling machine. These fears are real and understandable. Yet, despite centuries of technological disruption, we do not live in a world of massive unemployment. This is because, as old roles are erased, many more arise in their stead. But these fears keep recurring, because, at any given time, it is easy to imagine losing specific jobs but it requires superhuman imagination to forecast what new opportunities will be created by transformative new technologies. Nowadays, most of us do jobs that would be hard to explain to a 19th century worker.

The real workforce concerns should not be about the technology itself, but whether the economic system shares the benefits of new productivity fairly across society versus concentrating wealth only among the very richest \citep{Stiglitz} and whether machine learning systems are misused to magnify existing forms of inequality \citep{mathDestruction}. When displacements due to new technology occur, their effects can be eased by social safety nets and better educational foundations (for employment flexibility and retraining). Conversely, a society which fails to distribute wealth and economic gains fairly has much bigger problems than just the impact of AI.

Fears of new technology seem to be human nature. I suspect many people view the ``normal'' state of things as being how they were when they came of age, and they view any significant change as scary. Yet, nearly all of our familiar modern technologies were viewed as threatening by some previous generation.

The fear of human-created life has been with us for a long time. Notably, 18th-century scientists discovered electricity. As they searched to understand it, they discovered the life-like effect of galvanism, that the muscles of dead frogs could be stimulated by electrical currents. Had the secret of life been discovered? This  inspired Mary Shelley's novel \textit{Frankenstein; or The Modern Prometheus}, in which an ambitious university student uses modern science to create new life \citep{frankenstein}. Today, the story is vivid and evocative, but, intellectually, we recognize it as preposterous. The fear of AI is essentially the same irrational fear; SkyNet is Frankenstein's monster, but with neural networks as the Promethean spark instead of galvanism.\footnote{In fact, \textit{Frankenstein} is presented as a cautionary tale about the quest for knowledge in general. Victor Frankenstein tells his story as a warning when he learns that Captain Walton is himself driven by an obsessive quest for knowledge that is entirely unrelated to Frankenstein's.} At present, the Terminator's autonomous AI is only slightly more plausible than its ability to travel backwards in time.


\subsection{New AI  will be new tools for artists}

Some general trends around the evolution of technology and art seem quite robust. As discussed above, current AI algorithms are not autonomous creators, and will not be in the foreseeable future. They are still just tools, ready for artists to explore and exploit. New developments in these tools will be enthusiastically adopted by some artists, leading to exciting new forms and styles that we cannot currently foresee. Novices will have access to new simplified tools for expression. It is possible that some tasks performed by human artists will fade out, but these will generally be mechanical tasks that do not require much creativity because they fill societal functions other than artistic expression. Some traditional arts may fade simply due to seeming old-fashioned. This is the nature of art: nothing is fresh forever, which is not to be blamed on technology. 
 Artistic technology is a ``imagination amplifier'' \citep{imaginationAmp} and better technology will allow artists to see even further than before.

Aside from general trends, it is hard to make specific predictions about the art of the future. Les Paul, who invented the solid-body guitar in the 1940s, himself primarily performed light pop, country, and showtunes, and could hardly have predicted how the electric guitar would be used by, say, Led Zeppelin, just as it's hard to imagine Daguerre predicting Instagram.
More generally, making predictions about how AI technologies might transform society is very hard because we have so little understanding of what these technologies might actually be \citep{Brooks}. Even the science fiction writers of the 1950s and 1960s completely failed to imagine the transformative power of the Internet and mobile computing \citep{failedInternet}; for them, the computers of the future would still be room-scale monstrosities that one had to sit in front of to operate. But they did predict  moon colonies and replicants by 2018.

In short, we cannot predict what new inventions and ideas artists will come up with in the future, but we can predict that they will be amazing, and they will be amazing because they make use of technology in new, unpredictable ways.


\section{What is an Artist?}

So far, I have described how computer technologies are currently accepted as tools for artists, not as artists themselves. Why is this? After all, computers can do other human tasks like speak, search, print, navigate, and, to some extent, drive cars. There are several obvious reasons why computers don't make art, including tradition, the incentives involved, and the relatively predictable nature of existing automation. But, still one could imagine an alternate history in which some machines or computer programs had already been called artists. I believe there is a more fundamental reason, which explains not just why this has not happened, but why it is unlikely to happen anytime soon.

In this section, I theorize about what are the prerequisites for an entity being an artist, focusing on my hypothesis that art is a social behavior. I will then apply this idea to AI in the next section.
I also explore several alternative hypotheses for what could make an AI as an artist. Most of these alternative hypotheses identify some attribute of human artists, and then hypothesize that that attribute is needed for an AI to be an artist.



\subsection{Art is social}

\textit{``What an artist is trying to do for people is bring them closer to something, because of course art is about sharing: you wouldn't be an artist if you didn't want to share an experience, a thought.'' -- David Hockney} \citep{Hockney}

Why do we create and consume art? I argue that art is primarily a social behavior: art is about communication and displays between people. For example, people often speak of art as being about personal expression, which is an act of communication.

Any human can make art, because humans are social creatures. A painter cranking out the same conventional landscapes for tourists year after year is still considered an artist. A child’s drawing might only be interesting to their family, but it’s still art.

I am directly inspired by the theory, going back to Charles Darwin, that art-making is an adaptive product of our biological evolution. Dutton \cite{Dutton} sets out a persuasive argument for this theory, which I briefly summarize, though I cannot do it justice. Creating art served several functions for our Pleistocene ancestors. Art-making served as a fitness signal for mating and sexual selection.  Art can also be used as displays of wealth and status. Storytelling, music, and dance strengthen social bonds within a group. Storytelling additionally plays a very important role of communicating information that would otherwise be hard to share.

I observe that each of these functions of art is  \textit{social}: art arose as forms of communication, displays, and sharing between people.  Although art takes many different forms in different cultures today, each of these forms serves one or more of the same basic social functions that it did in the Pleistocene.

I generalize this theory beyond humans to hypothesize: \textit{art is an interaction between social agents.} A ``social agent'' is anything that has a status akin to  personhood; someone worthy of empathy and ethical consideration. Many of our other behaviors are interactions between social agents, such as gifts, conversation, and social relationships like friendship, competition, and romance.

In contrast, while we can get emotionally attached to our computers and other possessions, we feel no real empathy for their needs, and no ethical duty toward them.   
Possessions can participate only in shallow versions of these interactions. For example, we frequently talk about our possessions with statements like ``The brakes on my bicycle were complaining so I gave it a new pair as a gift and now it's much happier. I love my bike.'' This statement indicates emotional attachment but not true empathy with the bike's feelings, despite the anthropomorphizing language.
We don't live in a social hierarchy with our possessions: we do not compete with them for status, or try to impress them. We care about what other people have to say because we care about other people; we care about what computers have to say only insofar as it is useful to us.
We generally treat conversational agents (like Siri and Alexa) as user interfaces to software, not like people.

Art-making does also have non-social benefits to the artist. For example, art-making can help practice skills like dexterity and problem solving. Creating art is often pleasurable or meditative in itself.  But these benefits are secondary to their social benefits: they are not the reasons that evolution has produced art as a human activity. Similarly, one may also talk or sing to oneself while alone, but talking and singing are still fundamentally social activities.



Note that the evolutionary argument here is optional; one can discuss whether art is fundamentally social without it. But I believe that the evolutionary view of art gives some additional understanding.

\subsection{Non-Human Authors}

As we have seen, despite many technological advances, current algorithms are not accepted as artists.
There are a few other existing examples where objects are created by authors or processes that are not human-driven. These give some support to this theory. 



\mypara{Natural processes.} 
Natural processes, including landscapes like the Grand Canyon or the HuangShan Mountains, are not considered art, even though they may be extraordinarily beautiful and change one's perspective immensely. Beautiful structures  made instinctively by animals, such as honeycombs and coral, are not considered art. This indicates that simply creating complex and beautiful outputs is not itself sufficient for art, since there is no creative social communication in these cases. 

\mypara{Animals.}
Some higher mammals, including chimpanzees, elephants, and dolphins, have been trained to paint \citep{animalArt,Dutton}. Many writers are skeptical of animal-made art.
Typically, the animal's owner or handler steers the process, letting the animal throw paint on the canvas, then stopping the painting when they believe it is done, then selecting which works to show; the animals seem to show no interest in the artwork afterward. 
Animal artwork has not had any significant cultural impact or popularity; it seems to have been largely the product of media stunts. (People for the Ethical Treatment of Animals have recently tried to claim copyright in favor of a monkey, but failed \citep{monkeyselfie}, as US copyright law only allows humans to claim copyright.)  

The most interesting aspect of this discussion is not whether animals can create art, but \textit{how we decide.}
Discussions of whether animal-made artifacts are art are not based on \textit{a priori} rules whether animals can be artists. Instead, they are attempts to study the evidence of the animal's behavior around the artwork, and, from that, to infer whether the artwork is some form of inner expression \citep{animalArt}, or an artifact that the animal has a special appreciation for \citep{Dutton}.
In other words, we are open to the idea of animals creating art, because they can have social relationships with us. It's just that we haven't found any other creature that satisfies our criteria for creating art, whatever they are.

\begin{figure}
    \centering
    (a)
\includegraphics[width=2in]{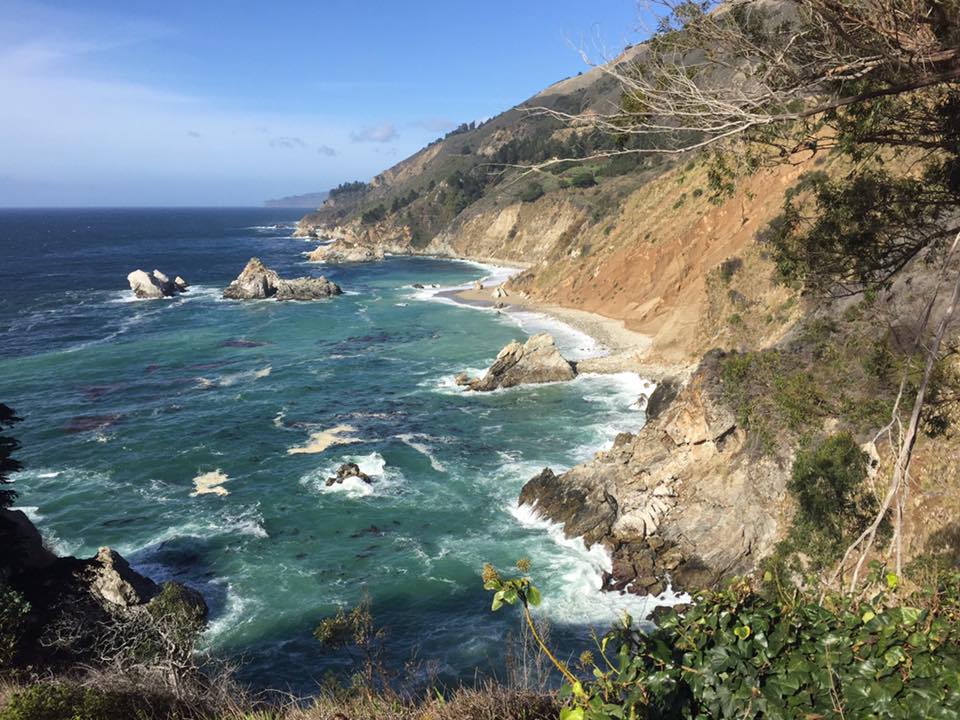}~~
(b) \includegraphics[height=1.5in]{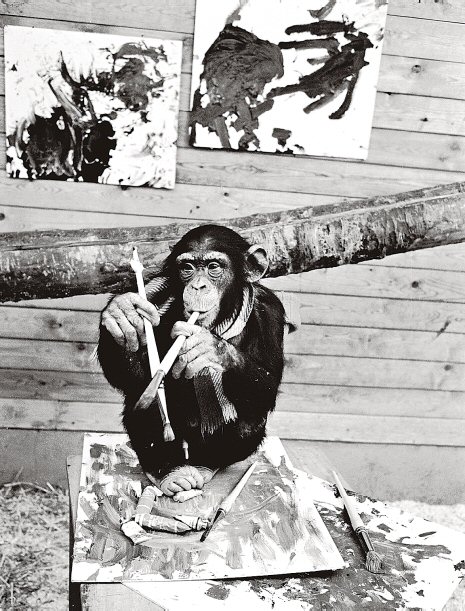}
    \caption{
    \label{fig:artmonkey}
  Not art.
   (a) Beautiful landscapes are not considered art.
(b)    Pierre Brassau, the monkey painter. He was part of a 1963 hoax that satirized modern art.
    }
\end{figure}


\subsection{Judging the work instead}

It is tempting to judge whether a computer is an artist based solely on the merits of the work that it produces. In this hypothesis, whether or not a computer can be an artist is a judgment of the quality of the work that it produces, independent of the properties of the computer itself.  If an algorithm outputs a continual stream of diverse, stimulating, beautiful, and/or skillful outputs, without many duds, we might be quite tempted to call this algorithm an artist. 
The better computer-generated art becomes, the more we will hear questions about whether computers are artists. 

Skill is clearly not the real requirement for someone or something to be an artist. Any human can make art, including unskilled amateurs and children. Conversely, computers can already be programmed to create infinite sequences of dazzling realistic or abstract imagery, exhibiting technical proficiency way beyond the typical human capacity.

I suspect that, when we look at a computer's output and ask ``is this work good enough to call the computer an artist?", we are not actually judging the quality of the work \textit{per se}. Instead, we are really looking for evidence that the system itself is intelligent, conscious, and feeling: traits that we associate with social agents. 
No matter how skillful and surprising a computer's output is, we will not accept it as an artist until we infer some sort of social being inside.

\subsection{An Intent Machine}

Another hypothesis is as follows: in the modern art world, the role of the artist is to supply the ``intent'' and the ``idea'' for the work; it is not necessary for the artist to execute on the work, other than coordinating in its production.  For humans, this is clearly true in numerous examples, such as those in Section \ref{sec:procedural}, as well as appropriation works like Duchamp's Readymades and Richard Prince's questionable Instagram reproductions; artists can also employ helpers or crowdworkers, such as in Scott Draves' \textit{Electric Sheep} and Aaron Koblin's \textit{Sheep Market} \citep{sheepmarket}. Consequently, for a computer to be an artist it simply needs to supply an intent.

It is easy to imagine designing a system that creates intent and even coordinates the labor of producing a work. For example, one could write a simple procedural algorithm to generate very basic intents (e.g., ``portray ominous landscape''), or sample intents from a Recursive Neural Network trained on artists' statements scraped from the web. Or the method could randomly select some news item, photograph, or historical event, and randomly sample some attitude toward that thing.  Starting from this intent, crowdworkers could be used to refine the idea and convert it into a new image, similar to systems like Soylent \citep{soylent}. One could also automate steps of the process, e.g., using GANs \citep{GANs} to generate entirely new starting images from scratch. Crowdworkers could also be used to rate and evaluate the outputs of the system, selecting just the best results and discarding the rest. Final steps of the process could also automatically hire professional designers, e.g., from sites like Upwork or 99Designs. This system could then run continuously, generating new images over time (with payment being automatically made to the crowdworkers involved). Workers could group images with common themes and intents, and create separate collections around these themes. Artist's statements could generated around these themes.  The system's preferences could grow and adapt over time as more data are gathered or external data streams (e.g., photography blogs) change.

Suppose someone were to build this system, calling it, say, The Intent Machine, and exhibits its work in an art show or gallery. Suppose, moreover, they convinced the curator or gallery owner to credit The Intent Machine with authorship of the works it had created, but fully disclosed the procedure by which it worked. Would people credit it as the artist who had authored its works, or would they say that the system-builder is the real artist here?

I believe that, in general, the consensus would be that the system-builder is the real artist here, and that this is really an artwork about probing the nature of computer-generated art, or the nature of the commercial artworld.  Note that the procedure used to define the system is not fundamentally different than any other procedural computer-generated art algorithm (as in Sections \ref{sec:procedural} and \ref{sec:NPR}).  Even if the work itself ended up being quite good, viewers would ask why it is good, and it is doubtful that the computer's own contribution would be judged as  significant beyond those of the humans involved.

Are artists just ``intent machines''? If art is a social act, then the answer is no.

\subsection{Creativity, Growth, Responsiveness, and so on}

Finally, three related criteria that have intuitive appeal are the notions of creativity, growth, and responsiveness. They are important for human artists, so perhaps they should be for AI artists as well.

Several authors have proposed energy terms or criteria for \textit{creativity} \citep{InnovationEngine,CAN}. These are often attempts to express the idea that the system's output should somehow ``surpass'' the human programmer and/or the training data.
For example, Colton \cite{ColtonCreativity} proposes that we judge the creativity of a system, in part, by whether the system's output \textit{surprises} the system's author. I believe that this is too weak a criterion, since many mechanical  or algorithmic phenomena may be surprising to their own discoverer or author at first. 
For example, the basic algorithm that produces Mandelbrot set images can be specified in a single sentence,\footnote{Color the image location $(x,y)$ proportional to the number of steps the iteration $z_{n+1} = z^2_n + x+yi$ requires to reach $|z|\ge\tau$, starting from $z_1=0$, where $\tau$ is a large constant.} 
yet produces dazzling animations of infinite complexity.\footnote{For example: \url{https://www.youtube.com/watch?v=PD2XgQOyCCk}}  The Mandelbrot set is very surprising and produces beautiful, unprecedented images, but we do not call its iteration equation creative, or an artist.  

All current procedural art systems, such as the Mandelbrot set, have a recognizable style, and, after awhile, lose their novelty. I believe that the same will be true for systems specifically designed with ``creativity'' objectives \citep{CAN,InnovationEngine}. Unlike human artists, these systems do not grow or evolve over time.

Perhaps an AI artist would need to exhibit some form of \textit{growth.}
For example, Harold Cohen, the fine-art painter who began, in 1968, to write software to generate art, described the evolution of his views: ``Ten years after [1968], I would have said `look, the program is doing this on its own.' ... Another ten years on and I would have said `the fact that the program is doing this on its own is the central issue here,' denoting my belief in the program's potential and growing autonomy over the whole business of art-making. ... It was producing complex images of a high quality and I could have had it go on forever without rewriting a single line of code. How much more autonomous than that can one get? ... [But] it's virtually impossible to imagine a human being in a similar position. The human artist is modified in the act of making art. For the program to have been similarly self-modifying would have required not merely that it be capable of assessing its own output but that it had its own modifiable worldview ...'' \citep{HaroldCohenSIGGRAPH}

In any existing system, it is easy to think of trivial ways for the system to evolve and change over time, e.g., subtly change the color palette or the training data over the years. Superficial growth is easy; meaningful growth is hard to even define for a computer AI. If someone could design a system that produces a sequence of art that is meaningful to people and also significantly evolves over time, that would be truly remarkable. Another missing piece from current systems is an artists' ability to \textit{respond} meaningfully to their culture, experiences, world events, responses to their work, and other aspects of their environment. It seems hard to imagine achieving these goals without enormous technological advances --- they may not be possible without true AI or social AI in some form.


\subsection{Definitions of Art}

For guidance about the nature of art, we could have also looked to existing definitions of art. 
However, there is no Royal Society that prescribes what is and what is not valid art. Instead, art is a phenomenon that results from the interplay of cultural institutions and the general population that we can analyze, and it changes over time.  
Philosophers  have attempted to  devise concise definitions of art that include all existing types (music, dance, painting, etc.) and styles of art. The Institutional Definition, originated in the 1960s in response to conceptual art, states, roughly, that art is anything in a style that is broadly accepted as art \citep{dantoArtworld,dickie1971,Levinson79}. A more fine-grained approach is to identify attributes common to many different types of art \citep{gaut2000,gaut2005,dutton2000,Dutton}. Each of these definitions of art is  an attempt to fit the data, and 
draw a line between those things that we call art (like theatre) and those that we do not (like spectator sports). 
Understandably, these definitions all assume that the artist is always human, without exploring much whether non-humans can create art, and thus do not provide much guidance for this discussion.

\subsection{Attribute theories in general}

Many of the theories in this section have the following recipe: identify some attributes of human artists, and then hypothesize that AIs with these attributes will be considered artists.  Artists make high-quality work; artists supply intent; artists are creative; artists grow.

The development of Cohen's thinking illustrates this. 
Cohen initially thought that a machine that makes high-quality art could be considered an artist.  The more he observed his own software, the more something seemed missing. He noted that human artists grow over time, and his system didn't.  If he'd added some form of ``growth'' to his system, would it then be an artist?

Maybe someday someone will develop a system with enough of these attributes to cross some intangible threshold and thus be perceived as an artist. However, it seems very hard to reason concretely about this possibility. How do we define any of these attributes precisely? How much is enough?

Furthermore, many of these attributes are not really required of human artists. 
Any human can make art, even if it is not very original or surprising; the artist need not grow noticeably or respond to culture or feedback. We do judge the work by these attributes, but there is no minimum requirement for humans to make art. 

In contrast, the social theory makes a much more concrete statement. Art is fundamentally a social interaction, and thus can only be made by social agents. The social theory has the additional appeal of being based on a plausible evolutionary hypothesis for the reasons we create art.

\section{Will an AI ever be an artist?}

With this background, I now turn to speculating about the future.
As we have seen, authorship of all current algorithmic art is assigned to the human author behind the algorithms. 
Will we ever say that an AI itself created art? Will we ever recognize a piece of software as the author of a work of art?  




\mypara{Human-Level AI.}
If we ever develop AI with human-level intelligence and consciousness, by definition, it would be able to  create art, since it would have the same capacity for consciousness, emotions, and social relationships. But, as discussed in Section \ref{sec:ai}, this scenario is  science fiction and we have no idea if this is possible or how it would be achieved. Making meaningful predictions about a world with ``true AI'' is impossible \citep{Brooks}, because we have so little idea of how specifically this AI would actually operate. 
Moreover, this AI would transform society so much as to make it unrecognizable to us. We may as well speculate about what kind of artwork is made by  aliens from outer-space --- if do we ever meet them, we will have more pressing questions than what kind of music they like.

Hence, the interesting question is whether there could be computer-authored artwork \textit{without} human-level intelligence or consciousness.


\mypara{Social AI.}
If, as I have argued, creating art is a fundamentally social act of expression and communication, then it follows that \textit{AI can be granted authorship when we view the AI as a social agent, and it is performing some communication or sharing through art.} 

What does it mean for us to view an AI as a social agent? We have to view the AI as deserving of empathy and ethical consideration in some way. However, the AI does not need human-level intelligence; just as we have social relationships with our pets. But we do expect that the AI has something to say socially, something that suggests an inner consciousness and feeling.

Short of true intelligence (the science fiction scenario), I think that the only way this can happen is through ``shallow AI'' agents.
People are sometimes ``fooled'' by shallow AI. The classic example is Eliza, a simple text-based “psychiatrist” program developed in 1964, based on simple pattern-matching and repetition of what the user types \citep{eliza}. It was meant as a demonstration of the superficiality of the AIs of the time, but, unexpectedly, many people attributed human-like emotions to the machine. Since then, there are many anecdotes of people being fooled by ``chatbots'' in online settings \citep{ArtificialTA,IvanaEpstein}, including the recent plague of Twitter bots \citep{TwitterBot}. But, once the veil is lifted, it is clear that these chatbots don't  exhibit real intelligence.  

Some software and robots have been designed to have relationships with their owners, including talking dolls, Tamagotchi, and Paro therapeutic baby seals.
A related effect is that people behave toward their computers as if they were social agents in certain ways \citep{TheMediaEquation}, even when they don't believe that they are intelligent.
For example, dialogue systems like Siri and Alexa all use female voices by default, based on many findings that  male and female users both respond better to female voices \citep{SiriFemale,TheMediaEquation}.  

Perhaps, for many users, the system doesn't need to be truly intelligent, it just has to be \textit{perceived} as a social agent, like a Siri or Alexa that you can ask to make you an artwork  \citep{AttnGAN}. 
The day may come in which these agents are so integrated into our daily lives that we forget that they are carefully-designed software. One can easily imagine the development of AI that  simulates emotion and affection toward the user; it is easy to imagine, for example, a toy doll that paints pictures for its owner along as one of many behaviors designed to display companionship and affection.

\mypara{Non-social AI.}
It seems possible that non-social algorithms could be successfully promoted as artists; there have been a few tentative forays in this direction, e.g., \citep{PaintingFool,CAN}. For the reasons given above, I am skeptical that such methods will be accepted as true artists without some plausible belief about their underlying social and/or conscious attributes.

Perhaps a curator at a well-known museum would download or otherwise acquire various artifacts from software ``artists,'' and list the software systems as the authors. There would be controversy, and discussion in newspapers and journals. Perhaps other curators and galleries would follow suit. Perhaps people would find enough value in these computer-generated artworks, while also being convinced that no human could be rightly given credit for their works. This sort of process has happened for things like abstract expressionism, and not for chimpanzee art. Could it happen for computer art?  

\mypara{Suitcase words.}
The term ``artist'' could come to be used as having multiple meanings, just as words like ``intelligence'' and ''learning'' have come to mean something different for humans than they do for algorithms \citep{Brooks}. A software program that, say, automatically stylizes your photos, could be called an ``artist'' in the same way that software applications like ``calendar'' and ``mail'' programs have replaced their physical-world namesakes.  Unfortunately, the use of the same word to mean different things in different contexts causes endless confusion, as discussed in Section \ref{sec:ai}.





\mypara{Dangers.}
A continual danger of new AI technology is that human users misunderstand the nature of the AI \citep{mathDestruction}. When we call a shallow AI an ``artist,'' we risk seriously misleading or lying to people. I believe that, if you convince people that an AI is an artist, then they will also falsely attribute emotions, feelings, and ethical weight to that AI.  If this is true, I would argue that calling such AIs ``artists'' is unethical. It leads to all sorts of dangers, including overselling the competence and abilities of the AI, to misleading people about the nature of art.

It seems likely that some companies will not have any scruples about this. For example, Hanson Robotics has promoted (in many contexts) a social robot as a truly intelligent being, even though it is clearly nothing more than a ``chatbot with a face'' \citep{Sophia} or, in Yann LeCun's words, a ``Potemkin AI'' \citep{SophiaYann}.

A related concern is that we deprive the AI's designer(s) of authorship credit.
At present, we credit the author of a piece of automatic software with the output of that software. This usually acknowledges the skill and effort required to engineer and iterate with software so that it produces good outputs. Artistic credit is important for understanding the real sources of how something was made.

Outside of science fiction, I can see no positive benefit to calling a computer an artist, but I do see dangers.

\section*{Conclusion}

I do not believe that any software system in our current understanding could be called an ``artist.''  Art is a social activity. I mean this as a warning against misleading oneself and others about the nature of art. Of course, the ambitious reader could take this as a challenge: I have laid out some of the serious objections that you must overcome if you wish to create a software ``artist.''  I don't think it can be done anytime soon, but I also know that proving critics wrong is one of the ways that art and science advance.

One of my main goals in this essay has been to highlight the degree to which technology contributes to art, rather than being antagonistic. 
We are lucky to be alive at a time when artists can explore ever-more-powerful tools.
Every time I see an artist create something wonderful with new technology, I get a little thrill: it feels like a new art form evolving. Danny Rosin's \textit{Wooden Mirror}, Jason Salavon's \textit{The Top Grossing Film of All Time, 1x1}, Bob Sabiston's \textit{Snack and Drink}, Michel Gondry's \textit{Like A Rolling Stone}, Kutiman's \textit{ThruYOU}, Amon Tobin's \textit{Permutation}, Ian Bogost's \textit{Cow Clicker}, Christian Marclay's video installations, \'{I}\~{n}igo Quilez's procedural renderings, and Wesley Allsbrook's and Goro Fujita's Virtual Reality paintings are a few examples of artworks that have affected me this way over the years. Today, through GitHub and Twitter, there is an extremely fast interplay between machine learning researchers and artists; it seems like, every day, we see new tinkerers and artists Tweeting their latest creative experiments with RNNs and GANs (e.g., @JanelleCShane, @helena, @christophrhesse, @quasimondo, @DrBeef\_).

Art maintains its vitality through continual innovation, and technology is one of the main engines of that innovation. Occasionally, the  avant garde has tremendous cultural impact: electronic music and sampling was once the domain of experimental electronic and musique concr\`{e}te pioneers, like Wendy Carlos and Delia Darbyshire. Likewise, at one time, computer-animated films could only be seen at obscure short-film festivals. Today, we are seeing many intriguing and beguiling experiments with AI techniques, and, as artists' tools, they will surely transform the way we think about art in thrilling and unpredictable ways.

\mypara{Acknowledgements.}  Thanks to Shira Katz, Alvy Ray Smith, Craig Kaplan, Shiry Ginosar, and Dani Oore for valuable comments on the manuscript. Thanks to everyone who shared discussion and/or encouragement online, including Aseem Agarwala, Mark Chen, Lyndie Chiou, Michael Cohen, James Landay, Nevena Lazic, Jason Salavon, Adrien Treuille, and many others.

\bibliographystyle{acm}
\bibliography{paper}



\end{document}